\documentclass[journal]{IEEEtran}

\usepackage{graphicx}
\usepackage{amsmath}
\usepackage{booktabs}
\usepackage{multirow}  
\usepackage{textcomp}
\usepackage{subfigure}
\usepackage{amssymb}
\usepackage{cite}
\usepackage{hyperref}
\usepackage[switch]{lineno}
\usepackage{color}
\definecolor{mygreen}{RGB}{50,205,90} 
\definecolor{myyellow}{RGB}{238,173,14} 
\usepackage{makecell}

\usepackage{colortbl}
\usepackage{color}
\usepackage[table,xcdraw]{xcolor}

\ifCLASSINFOpdf
\else
\fi
\begin{document}
%
\title{Image Captioning via Dynamic Path Customization}
%
%
%


\author{Yiwei Ma*,
        Jiayi Ji*,
        Xiaoshuai Sun$^\dag$,~\IEEEmembership{Member,~IEEE}, 
        Yiyi Zhou,~\IEEEmembership{Member,~IEEE}, \\
        Xiaopeng Hong~\IEEEmembership{Member,~IEEE},
        Yongjian Wu, 
        Rongrong Ji,~\IEEEmembership{Senior Member,~IEEE}
\IEEEcompsocitemizethanks{
\IEEEcompsocthanksitem $^*$Equal Contribution. $^\dag$Corresponding Author. 
\IEEEcompsocthanksitem J. Ji, R. Ji, X. Sun (e-mail: xssun@xmu.edu.cn), Y. Zhou and Y. Ma are with Key Laboratory of Multimedia Trusted Perception and Efficient Computing, Ministry of Education of China, Xiamen University, 361005, P.R. China..\protect
\IEEEcompsocthanksitem X. Hong is with School of Computer Science and Technology, Harbin Institute of Technology, Harbin, 150006, China.\protect
\IEEEcompsocthanksitem Y. Wu is with Youtu Laboratory, Tencent, Shanghai 200233, China.\protect
}
\thanks{
This work was supported by National Key R\&D Program of China (No.2023YFB4502804), the National Science Fund for Distinguished Young Scholars (No.62025603), the National Natural Science Foundation of China (No. U21B2037, No. U22B2051, No. 62072389), the National Natural Science Fund for Young Scholars of China (No. 62302411), China Postdoctoral Science Foundation (No. 2023M732948), and the Natural Science Foundation of Fujian Province of China (No.2021J01002,  No.2022J06001).}
}

%
%

\markboth{IEEE Transactions on Neural Networks and Learning Systems,~Vol.~xx, No.~xx, August~xxxx}%
{Shell \MakeLowercase{\textit{et al.}}: Bare Demo of IEEEtran.cls for IEEE Journals}
%



\maketitle

\begin{abstract}
This paper explores a novel dynamic network for vision and language tasks, where the inferring structure is customized on the fly for different inputs. Most previous state-of-the-art approaches are static and hand-crafted networks, which not only heavily rely on expert knowledge, but also ignore the semantic diversity of input samples, therefore resulting in sub-optimal performance. To address these issues, we propose a novel \emph{\textbf{D}ynamic \textbf{T}ransformer \textbf{Net}work} (DTNet) for image captioning, which dynamically assigns customized paths to different samples, leading to discriminative yet accurate captions. Specifically, to build a rich routing space and improve routing efficiency, we introduce five types of basic cells and group them into two separate routing spaces according to their operating domains, \emph{i.e.,} spatial and channel. Then, we design a \emph{\textbf{S}patial-\textbf{C}hannel \textbf{J}oint \textbf{R}outer} (SCJR), which endows the model with the capability of path customization based on both spatial and channel information of the input sample. To validate the effectiveness of our proposed DTNet, we conduct extensive experiments on the MS-COCO dataset and achieve new state-of-the-art performance on both the Karpathy split and the online test server. \color{black}{The source code is
publicly available at \url{https://github.com/xmu-xiaoma666/DTNet}}
\end{abstract}


%
\IEEEpeerreviewmaketitle

\section{Introduction}
\label{sec:intro}
%
%
%
%

\IEEEPARstart{I}{mage} captioning, which aims to generate a natural-language sentence to describe the given image, is one of the most fundamental yet challenging tasks in vision and language (V\&L) research. Recent years have witnessed its rapid development, which is supported by a series of innovative methods \cite{ye2018attentive,yu2018topic,huang2020image,zha2019context,park2018towards,yang2020auto,wu2020fine,ma2023towards}.

\begin{figure}
\centering 
  \includegraphics[width=1.00\columnwidth]{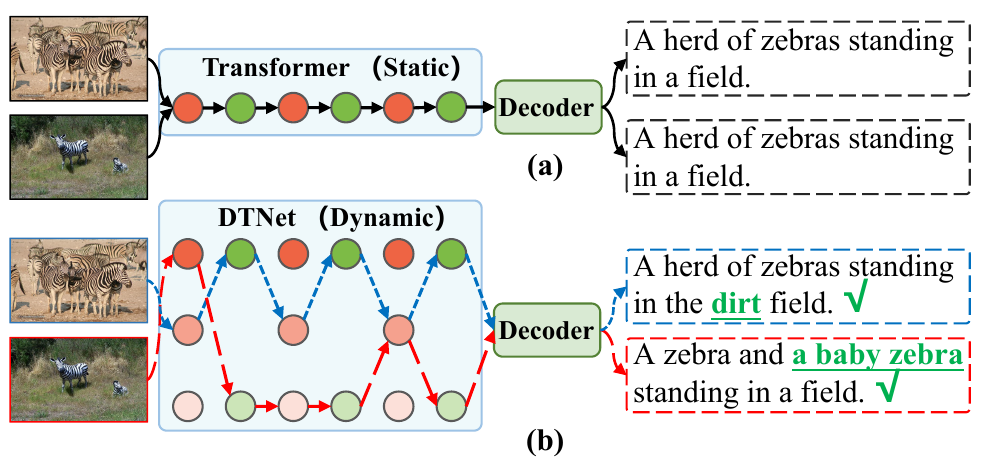}
  \vspace{-0.0cm}
  \caption{
        Illustration of Vanilla Transformer (static) and our DTNet (dynamic).  Circles of different colors represent different cells, and arrows of different colors represent data flows of different input samples. Note that orange and green circles are for spatial and channel operations, respectively. In this example, the static model (a) tends to generate the same sentence for similar images, while the dynamic network (b) can generate informative captions through dynamic routing. More examples are shown in Fig. \ref{fig:fig5}.
        }
  \label{fig:fig1}
\end{figure}

However,  most recent architectures \cite{li2019know,yang2018multitask,ji2022knowing,ma2022knowing,zhang2022adaptive,shao2022region,chaturvedi2020mimic,ji2022knowing} for image captioning are static,  where all input samples go through the same path despite their significant appearance difference and semantic diversity. There are two limitations to such static architectures: 1) The static network cannot adjust its architecture based on the input samples, therefore lacking flexibility and discriminability. As shown in Fig. \ref{fig:fig1} (a), due to the limitation of model capacity, when fed with semantically similar images, the static model tends to ignore the details and generates the same sentence, which has also been mentioned in previous works \cite{luo2018discriminability,wang2020compare,ma2022knowing}. Notably, such a ``safe'' captioning mode with static networks seriously prohibits generating informative and descriptive sentences for images. 2) The design of such static networks heavily relies on expert knowledge and empirical feedback from both developers and users.  

To address these issues, as illustrated in Fig. \ref{fig:fig1} (b), we explore a new paradigm to incorporate dynamic routing within the network design for adaptive and flexible captioning. However, three problems arise when applying typical dynamic routing strategies to image captioning: 
1) Most dynamic networks \cite{yang2019condconv,chen2020dynamic,zhang2020dynet} mainly focus on the dynamic design of convolution kernels, which ignores spatial multi-scale modeling and channel-wise modeling. 
2) Current dynamic methods place all candidate modules in the same routing space, resulting in low routing efficiency. 
3) Most routers in dynamic networks \cite{chen2021dynamic,zhang2020dynet,li2021revisiting,yang2019condconv,chen2020dynamic,zhou2021trar} are based on the Squeeze-and-Excitation \cite{hu2018squeeze} architecture, where spatial information is damaged by the \emph{Global Average Pooling} operation.
In this paper, we propose a novel input-dependent transformer architecture, dubbed as \emph{\textbf{D}ynamic \textbf{T}ransformer \textbf{Net}work} (DTNet), to solve all these three issues simultaneously. 
To address the first dynamic design issue, we introduce five basic cells to model input samples in both spatial and channel domains, thus building a richer routing space.
To address the second routing efficiency issue, we group five proposed cells into two separate routing spaces, which reduces the difficulty of routing optimization. Specifically, in the spatial domain, three cells are used for global, local, and axial modeling; in the channel domain, two cells conduct channel-wise modeling by projection and attention mechanism, respectively. 
To solve the last information damage problem, we propose a novel \emph{\textbf{S}patial-\textbf{C}hannel \textbf{J}oint \textbf{R}outer} (SCJR), which fully models both spatial and channel information of input samples to generate adaptive path weights. In particular, SCJR decouples the modeling of spatial and channel domains in two branches, and then the outputs from both branches are comprehensively processed to generate the appropriate path weights. 

Based on the aforementioned novel designs, during inference, different samples go through different paths adaptively for customized processing in DTNet. Note that most proposed basic cells are lightweight compared with Self-Attention and Feed-Forward Network, so our proposed DTNet achieves significant performance gains with negligible parameter increase over vanilla Transformer (\emph{i.e.,} 36.15 M \emph{vs.} 33.57 M).

In sum, our contributions are three-fold as follows:
\vspace{-0.0cm}
\begin{itemize}
    \setlength{\itemsep}{1pt}
    \setlength{\parsep}{1pt}
    \setlength{\parskip}{1pt}

    \item We propose an adaptive \emph{Dynamic Transformer Network} (DTNet) for input-sensitive image captioning, which not only generates more discriminative captions for similar images but also provides an innovative paradigm for diverse image captioning.
    
    \item We introduce five basic cells, which models input features with different mechanisms in the spatial and channel domain, to build a rich routing space for more flexible dynamic routing.
    
    \item We propose \emph{Spatial-Channel Joint Router} (SCJR), conducting dynamic path customization by joint consideration of spatial and channel modeling, to compensate for the information damage of previous routers.



    
\end{itemize}

\vspace{-0.0cm}
Extensive experiments on the MS-COCO benchmark demonstrate that our proposed DTNet outperforms previous SOTA methods by a considerable margin. Besides, the experimental results on the Flickr8K \cite{hodosh2013framing} and Flickr30K \cite{young2014image} datasets also validate the effectiveness and generalization of the DTNet.

\begin{figure*}
\vspace{-0.0cm}
\centering 
  \includegraphics[width=2.00\columnwidth]{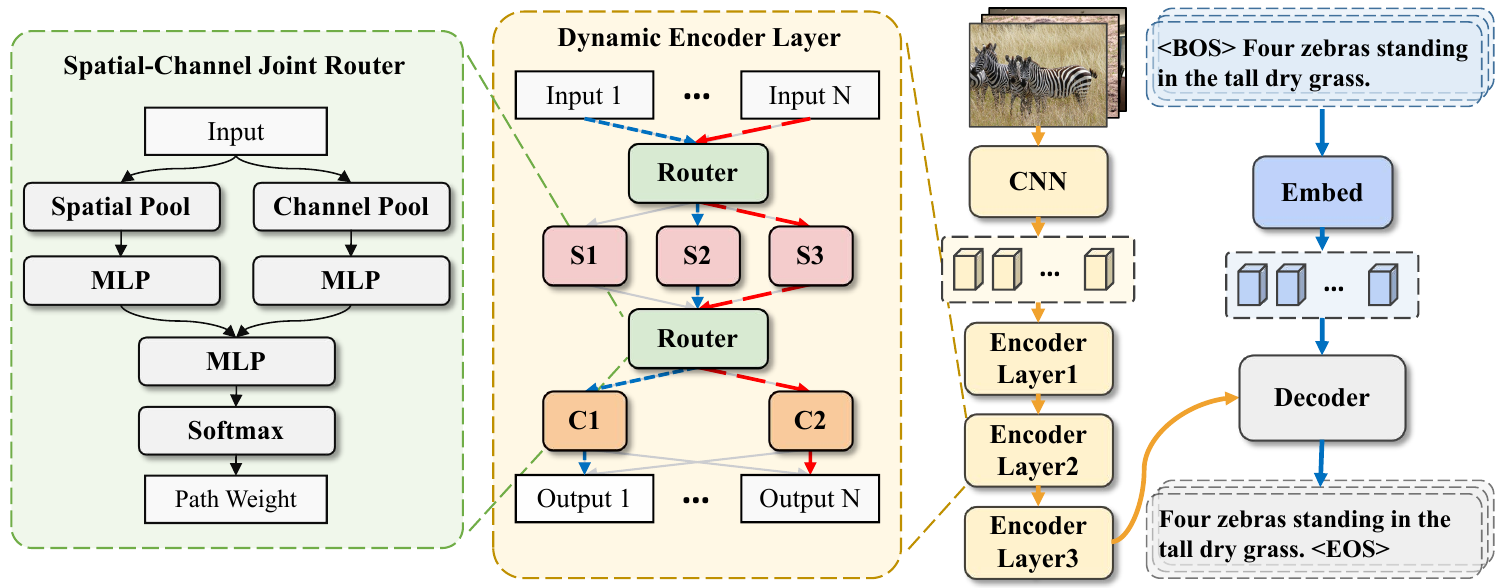}
  \vspace{-0.0cm}
  \caption{
        The framework of the proposed \emph{Dynamic Transformer Network} (DTNet)  for image captioning. The visual features are extracted according to \cite{jiang2020defense}. Next, stacked dynamic encoder layers are leveraged to encode the visual features with various input-dependent architectures, which are determined by our proposed \emph{Spatial-Channel Joint Router} (SCJR). Finally, the features from the encoder will be fed into the decoder to generate captions word by word. Residual connections in the encoder are omitted for simplicity. Best viewed in color.
  }
  \vspace{-0.0cm}
  \label{fig:fig2}
\end{figure*}

\section{Related Work}
Previous V\&L researches mainly focused on the design of task-oriented network architectures, which heavily depend on expert experience and empirical feedback. Unlike previous works, our proposed DTNet will dynamically customize the most suitable path for each input sample, which has seldom been explored in image captioning. In this section, we will first retrospect the development of image captioning and then give an introduction to the recent trends on dynamic networks.

\subsection{Image Captioning}
Image captioning is a challenging and fundamental task that promotes the development of multiple applications, \emph{e.g.,} human-computer interaction. With the rapid development of deep learning,  a great improvement can be observed with a flurry of methods \cite{lu2017knowing,yao2018exploring,jiang2018recurrent,gu2018stack,chen2019improving,herdade2019image,yang2019auto,pan2020x,luo2021dual,zhang2021rstnet,zhou2020unified,wang2021simvlm,zhang2021vinvl,lin2021m6,ma2022x}, \emph{e.g.,} SCST  \cite{rennie2017self}, Up-Down  \cite{anderson2018bottom},   AoANet  \cite{huang2019attention}, $M^2$Transformer  \cite{cornia2020meshed}, X-LAN  \cite{pan2020x} and OSCAR \cite{li2020oscar}. Generally, current captioning approaches for which may be classified into three types, \emph{i.e.}, template-based methods \cite{elliott2013image,mitchell2012midge}, retrieval-based methods \cite{ma2023beat,devlin2015exploring,karpathy2014deep,ma2022x}, and generation-based methods \cite{vinyals2015show,anderson2018bottom}. The template-based approaches  \cite{elliott2013image,mitchell2012midge} recognize visual concepts such as objects, attributes, and relationships and then insert them into predetermined phrase templates with several vacant slots to complete the captions. Template-based approaches can create grammatically accurate captions. However, the template is pre-defined, so the flexibility of language and the length of generated captions are severely constrained. The retrieval-based approaches \cite{devlin2015exploring,karpathy2014deep} try to search for the sentences that match the query images from the existing captions pool. Since these methods will not generate new captions to describe the given image, it is difficult for them to capture the uniqueness and complex semantics of the images. With the rise of generative models in natural language processing (NLP) and computer vision (CV), generation-based methods \cite{vinyals2015show,anderson2018bottom} are becoming the mainstream approaches for image captioning. Specifically, most generation-based methods follow the encoder-decoder paradigm, where the encoder is used to encode the image into visual vectorial representations, and the decoder is adopted to generate captions to describe the given images based on these vectorial representations. Due to their high flexibility and high performance, generation-based methods have been invested a lot of time and energy by researchers.

However, most previous models for image captioning are static, which heavily depend on professional design and hinders the generation of diverse sentences. Compared with static models, our DTNet conducts path customization based on input samples, therefore improving the flexibility and adaptability of the model. Moreover, a static model can only generate a single sentence for one image, while our DTNet can produce diverse sentences for the same input by controlling the path weights.

\subsection{Dynamic Network}

Empirical evidence in neurosciences \cite{Walther2011SimpleLD,LEE2020127} indicates that when processing different information, different parts of the hippocampus will be activated, which reveals the dynamic characteristic of the brain. Motivated by this finding, the dynamic network, which aims to adjust the architecture to the corresponding input, has become a new research focus in computer vision, \emph{e.g.,} image classification \cite{chen2020dynamic,yang2019condconv,yang2020resolution,huang2017multi,wang2023beyond}, object detection \cite{zhu2020dynamic,yang2023semi}, semantic segmentation \cite{liu2023rotated,li2020learning,wu20233d}, long-tailed classification \cite{duggal2020elf}. Chen \emph{et al.} \cite{chen2020dynamic} presented dynamic convolution, which is a new design that increases the complexity of the model without increasing the depth or width of the network. Li \emph{et al.} \cite{li2020learning} studied a new method, \emph{i.e.,} dynamic routing, to alleviate the scale variance in semantic representation, which generates data-related routes according to the scale distribution of images. Duggal \emph{et al.} \cite{duggal2020elf} proposed EarLy-exiting Framework (ELF) to address the long-tailed problem, where easy examples will exit the model first and hard examples will be processed by more modules. In the V\&L domain, Zhou \emph{et al.} \cite{zhou2021trar} proposed a dynamic design to capture both local and global information for visual question answering (VQA) by receptive field masking.


However, dynamic routing has seldom been explored for more general V\&L tasks, \emph{e.g.,} image captioning.  Directly incorporating existing dynamic mechanisms within the image captioning model will lead to sub-optimal performance. Thus, in this paper, we explore a dynamic scheme for image captioning to achieve better performance and generate diverse captions. It is worth noting that although TRAR \cite{zhou2021trar} also draws on the concept of the dynamic network, our proposed DTNet is quite different from it. Firstly, TRAR focuses on dynamic spatial modeling, so the dynamic idea is only reflected in the use of the dynamic receptive field, while the dynamic idea of our proposed DTNet is reflected in spatial and channel modeling at the same time. Secondly, TRAR is a Transformer with dynamic receptive field, which uses the attention mask to control the receptive field. Our DTNet proposes several modeling cells and the Spatial-Channel Joint Router to realize the input-sensitive network architectures. 
{\color{black}{
It is worth noting that our research introduces five novel basic cells, each having a unique role and contributing to the feature extraction process in a distinctive manner. The perceived marginal gains when considering the cells in isolation obscures the synergic performance gain we observed when combining them all together. It is the comprehensive methodology facilitated by this set of cells, rather than the individual performances, that really brings about the advancement in state-of-the-art that we have achieved.
}}

\section{Approach}

In this section, we present the details of the proposed \emph{Dynamic Transformer Network} (DTNet) for image captioning, where the specific network architectures vary with input samples. In particular, we first introduce the overview of DTNet in Sec. \ref{sec:overview}.  Then, we detail the architectures of five basic cells in the spatial and channel routing space in Sec. \ref{sec:spatial} and Sec. \ref{sec:channel}. Afterward, we show the design of our proposed \emph{Spatial-Channel Joint Router} (SCJR) in Sec. \ref{sec:router}. Finally, we elaborate on the objectives during training for image captioning in Sec. \ref{sec:objectives}.

\subsection{Overview}
\label{sec:overview}

Fig. \ref{fig:fig2} illustrates the overall architecture of our proposed DTNet. Given an image $I$, we first extract visual features $V \in \mathbb{R} ^ {H \times W \times C}$ following \cite{jiang2020defense}, where $H$, $W$, $C$ represent the height, width and channel dimension of the visual features, respectively.

Then, we feed the visual features into the proposed dynamic encoder to obtain the encoded visual features $\hat{V} \in \mathbb{R} ^ {H \times W \times C}$, which is formulated as:
\begin{equation}
    \hat{V} =  \eta(V),
\label{eq:Encoder}
\end{equation}
where $\eta(\cdot)$ denotes the operation in the dynamic encoder. As shown in the middle part in Fig. \ref{fig:fig2}, the forward paths are not static but adaptively determined by our proposed router, \emph{i.e.,}  the architectures vary with the inputs. 

In particular, the dynamic routing operation can be formulated as follows:
\begin{equation}
    \hat{Y} = \sum _{k=1} ^K \pi_k(x) Y_k,
\label{eq:Dynamic}
\end{equation}
where $K$ is the number of cells in the routing space, \emph{i.e.,} the number of candidate paths, $x$ is the input, $\pi_k(x)$ is the path weight for the $k$-th cell given $x$, $Y_k$ is the output of the $k$-th cell, and $\hat{Y}$ is the dynamic output.

Finally, the encoded visual features will be fed into the decoder, which follows the architecture of the Vanilla Transformer \cite{vaswani2017attention}, to generate the corresponding captions.

\begin{figure*}
\vspace{-0.0cm}
\centering 
  \includegraphics[width=2.00\columnwidth]{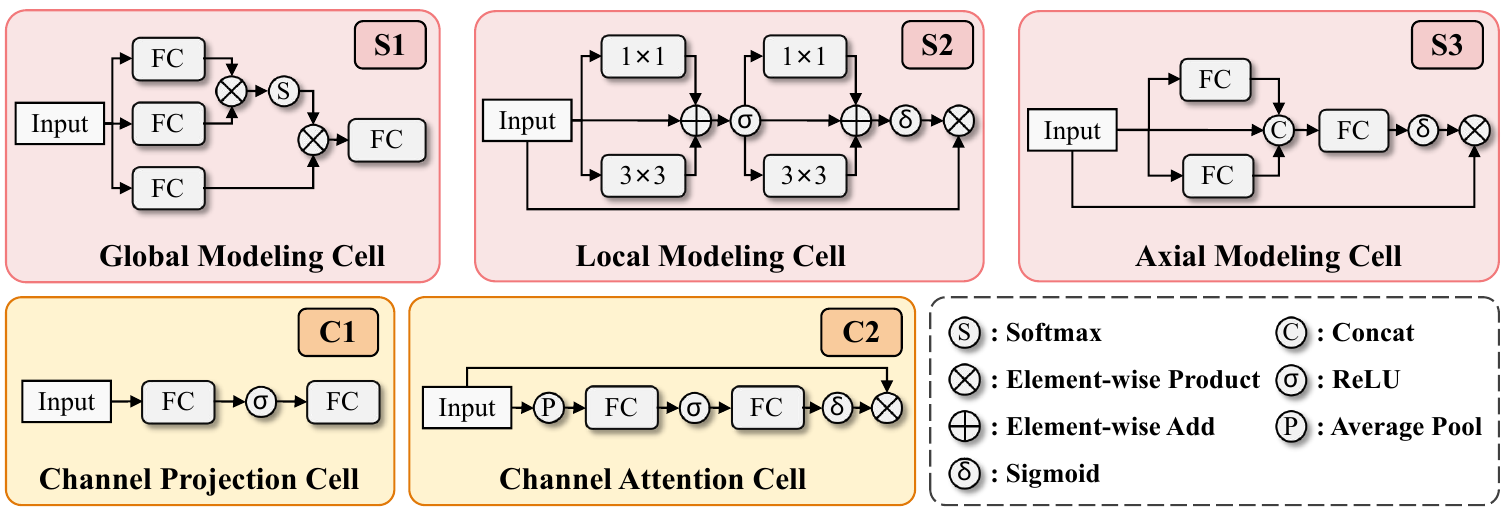}
  \vspace{-0.0cm}
  \caption{ 
     The detailed architectures of different cells in the spatial and channel routing space. BatchNorm is omitted for simplicity.
  }
  \vspace{-0.0cm}
  \label{fig:fig3}
\end{figure*}

\begin{figure}
\centering 
  \includegraphics[width=1.00\columnwidth]{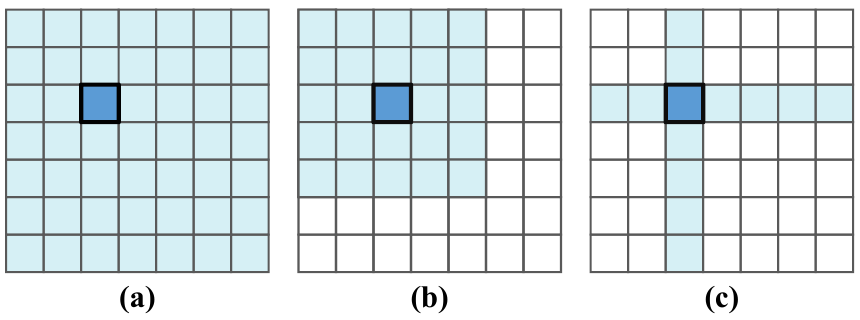}
  \vspace{-0.0cm}
  \caption{ 
      Receptive field illustration of different cells. (a) Global Modeling Cell, (b) Local Modeling Cell, (c) Axial Modeling Cell. The dark blue grid is the query grid, the light blue area is the receptive field, and the rest white area is the imperceptible area.
  }
  \vspace{-0.0cm}
  \label{fig:fig4}
\end{figure}

\subsection{Spatial Modeling Cells}
\label{sec:spatial}

To perceive the information of different receptive fields in the spatial domain, we tailor three cells, including \emph{Global Modeling Cell} (GMC), \emph{Local Modeling Cell} (LMC), and \emph{Axial Modeling Cell} (AMC), which are illustrated in the pink blocks in Fig. \ref{fig:fig3}. {\color{black}{Specifically, the GMC, LMC, and AMC have specific roles in modeling global, local, and axial information in the spatial dimension, respectively.}}

\subsubsection{Global Modeling Cell (GMC)}
To capture the global dependencies in the visual features, the global modeling cell (GMC) is introduced. As shown in Fig. \ref{fig:fig3} [S1], it is implemented with the multi-head self-attention (MHSA) mechanism of Transformer \cite{vaswani2017attention}.  

The $i$-th head of MHSA can be formulated as:
\begin{equation}
    h_i =  Softmax\left(\frac{\big(XW^Q_i\big) \big(XW^K_i\big)^\top}{\sqrt{d_k}}\right)\big (XW^V_i\big),
\label{eq:att}
\end{equation}
where $W^Q_i$, $W^K_i$, $W^V_i \in \mathbb{R}^{C \times C/\mathcal{H}}$ are learnable projection matrices, $\mathcal{H}$ represents the number of heads, $d_k$ is the number of channel dimension in $X W^K_i$. Thereafter, the  outputs of all heads are concatenated together as follows:
\begin{equation}
    MHSA(X)=\left[h_1;\dots;h_\mathcal{H}\right]W^O +X,
\label{eq:MHSA}
\end{equation}
where $[ \; ; \; ]$ is the concatenation operation across the channel dimension, $W^O \in \mathbb{R}^{C \times C}$  is the learnable parameter matrix. The receptive field of GMC is illustrated in Fig \ref{fig:fig4} (a).

\subsubsection{Local Modeling Cell (LMC)}
A series of works \cite{wei2021aligning,liu2021swin,dai2021coatnet,dong2021cswin,peng2021conformer,mehta2021mobilevit} demonstrate that translation invariance and local perception are critical for image recognition. Thus, in addition to global modeling, we further introduce LMC to perceive objects of different scales. As shown in Fig. \ref{fig:fig3} [S2], the LMC consists of two multi-branch convolutions, an activation function (\emph{i.e.,} ReLU) and a normalization function (\emph{i.e.,} Sigmoid). Each multi-branch convolution can be formulated as:
\begin{equation}
    X_{i+1}=
    BN_i(X_{i})+
    BN_i\left(F^{1 \times 1}_i(X_{i})\right)+
    BN_i\left(F^{3 \times 3}_i(X_{i})\right),
\label{eq:cnn}
\end{equation}
where $i \in \{0,1\}$ is the index of the multi-branch convolutions, $BN_i(\cdot)$, $F^{1 \times 1}_i(\cdot)$, $F^{3 \times 3}_i(\cdot)$ denote Batch Normalization \cite{ioffe2015batch}, $1 \times 1$ Conv and $3 \times 3$ Conv \footnote{\emph{$3 \times 3$} Conv is implemented by sequential convolutions with kernel sizes of $1 \times 1$  and $3 \times 3$. }, respectively. A ReLU activation module is used to connect these two multi-branch convolutions.

Afterward, we will normalize the output and apply the normalized weight to the input:
\begin{equation}
    Y=\delta(X_{2}) \otimes X_0,
\label{eq:cnn2}
\end{equation}
where $\delta(\cdot)$ is Sigmoid, $\otimes$ is element-wise multiplication. The receptive field of LMC is illustrated in Fig \ref{fig:fig4} (b).

\subsubsection{Axial Modeling Cell (AMC)}
Previous works \cite{dong2021cswin,tang2021sparse} have demonstrated that axial modeling in the image is critical for information perception. 
 Thus, we also introduce a simple cell to execute axial attention in the image, which is detailed in Fig. \ref{fig:fig3} [S3].
 
 Specifically, $X \in \mathbb{R}^{H \times W \times C}$ denotes the input of AMC.  We adopt two fully connected (FC) layers to over the width and height dimension of the input to obtain $X_W \in \mathbb{R}^{H \times W \times C}$ and $X_H \in \mathbb{R}^{H \times W \times C}$, respectively. Afterward, $X$ will be concatenated with $X_H$ and $X_W$ as follows:
 \begin{equation}
    X_{con}=[X;X_H;X_W], X_{con} \in \mathbb{R}^{H \times W \times 3C}.
\label{eq:concat1}
\end{equation}
 
For post-processing, an FC layer is used to reduce the channel dimension of $X_{con}$, which is followed by a Sigmoid function to normalize the output to get the axial attention weight. Finally, the input will be reweighted according to the attention weight, which can be formulated as:
 \begin{equation}
    Y=\delta\left(X_{con} W_{rec}\right) \otimes X ,
\label{eq:concat2}
\end{equation}
where $W_{rec} \in \mathbb{R}^{3C \times C}$ is the learnable parameter matrix. The receptive field of AMC is illustrated in Fig \ref{fig:fig4} (c).

\subsection{Channel Modeling Cells}
\label{sec:channel}

We explore two alternatives to model information in the channel domain, \emph{i.e.,} the projection-based and attention-based method. Specifically, we introduce two cells to model information through projection and attention. {\color{black}{The Channel Projection Cell (CPC) and Channel Attention Cell (CAC) operate in the channel dimension and perform different operations, respectively.}}

\subsubsection{Channel Projection Cell (CPC)}
CPC is a  projection-based method to model information in the channel domain, which is implemented with Feed-Forward Network (FFN) \cite{vaswani2017attention}. Concretely, as shown in Fig. \ref{fig:fig3} [C1], it consists of two FC layers with a ReLU activation in between: 
 \begin{equation}
    \operatorname{CPC}(X)=\sigma \left(X W^{CPC}_{1}+b_{1}\right) W^{CPC}_{2}+b_{2} ,
\label{eq:cpc}
\end{equation}
where $W^{CPC}_1 \in \mathbb{R}^{C \times 4C}$ and $W^{CPC}_2 \in \mathbb{R}^{4C \times C}$ are learnable projection matrices, $b_1$ and $b_2$ are bias terms, $\sigma(\cdot)$ is the activation function \emph{i.e.,} ReLU \cite{nair2010rectified}.

\subsubsection{Channel Attention Cell (CAC)}
CAC is a attention-based method for channel modeling, which is illustrated in Fig. \ref{fig:fig3} [C2]. Specifically, we adopt widely used Squeeze-and-Excitation (SE) \cite{hu2018squeeze} to implement it, which consists of a Multi-Layer Perceptron and a Sigmoid function as follows:
 \begin{equation}
    \operatorname{CAC}(X)=\delta\Big(\sigma \left( Pool(X) W^{CAC}_{1}\right) W^{CAC}_{2}\Big) \otimes X,
\label{eq:se}
\end{equation}
where $Pool(\cdot)$ is the average pooling operation in the spatial domain, $W^{CAC}_{1} \in \mathbb{R}^{C \times \frac{C}{16}}$ and $W^{CAC}_{2} \in \mathbb{R}^{\frac{C}{16} \times C}$ are learnable projection matrices, $\delta(\cdot)$ is the Sigmoid function, $\sigma(\cdot)$ is the ReLU activation function. 

{\color{black}{Specifically, the primary motivation behind integrating the CAC into our model stems from its crucial role in enhancing the representation capacity. By adjusting adaptive weights, the CAC selectively emphasizes and strengthens the most relevant feature channels. Through the channel attention mechanism, our model gains the ability to dynamically allocate attention to specific feature channels. This dynamic allocation enables the model to focus on the most informative channels while suppressing the less useful ones.
Furthermore, the inclusion of the CAC is designed to complement the Channel Projection Cell (CPC) within our model architecture. While the CPC is responsible for learning complex feature representations using stacked fully connected layers with non-linear activations, the CAC operates at a more granular level by fine-tuning the importance of individual feature channels. The combination of the CAC and the CPC results in a more powerful and flexible feature representation capability, as evident from the analysis of the last three rows in Tab.~\ref{tab:channel}.}}

\subsection{Spatial-Channel Joint Router}
\label{sec:router}
Most routers in previous dynamic networks \cite{zhou2021trar,chen2021dynamic,chen2020dynamic}  are based on the SE \cite{hu2018squeeze}, which corrupts the  spatial position information during global pooling.  To overcome this limitation, we propose a novel \emph{Spatial-Channel Joint Router} (SCJR), which is illustrated in the green block of Fig. \ref{fig:fig2}. In our proposal, the input features are processed by two branches, \emph{i.e.,} one for channel domain and the other for spatial domain. In the channel branch, the input is first squeezed in the spatial domain by Global Spatial Pooling ($\mathcal{GSP}$), and then processed by a multi-layer perceptron ($MLP$), which is formulated as:
\begin{align}
    \hat{X}_c =&  \sigma \Big( \mathcal{GSP}(X) W^{Cha}_{1} \Big) W^{Cha}_{2}, \\
    \mathcal{GSP}(X) =&\frac{1}{H \times W} \sum_{i=1}^{H} \sum_{j=1}^{W} X[i, j,:],
\label{eq:ChannelRouter}
\end{align}
where  $\sigma(\cdot)$ is the ReLU activation, $W^{Cha}_{1} \in \mathbb{R}^{C \times \frac{C}{r_1}}$, $W^{Cha}_{2} \in \mathbb{R}^{\frac{C}{r_1} \times p}$ ($r_1 = 16$ is the default setting in our experiment), $p$ is the number of candidate paths.

Similarly, the spatial branch can be formulated as:
\begin{align}
    \hat{X}_s =&  \sigma\Big( \mathcal{GCP}(X) W^{Spa}_{1} \Big) W^{Spa}_{2},\\
    \mathcal{GCP}(X) =&\frac{1}{C} \sum_{k=1}^{C}  X[:,:,k],
\label{eq:SpatialRouter}
\end{align}
where  $\mathcal{GCP}(\cdot)$ is the Global Channel Pooling, $W^{Spa}_{1} \in \mathbb{R}^{N \times N/r_2}$, $W^{Spa}_{2} \in \mathbb{R}^{N/r_2 \times p}$ ($r_2 = 7$ is the default setting  in our experiment), reshape operation is omitted for simplicity, $N$ is the number of grids, \emph{i.e.,} $N = H \times W$.

Finally, the outputs from the channel and spatial branches will be concatenated, and then fed into an MLP followed by the $Softmax$ normalization:
\begin{equation}
    \hat{W} =  Softmax \Big( \sigma\big( [\hat{X}_c;\hat{X}_s] W^{Joint}_{1} \big) W^{Joint}_{2} \Big),
\label{eq:Router}
\end{equation}
where $[\;  ; \;  ]$ is the concatenation operation of tensors, $W^{Joint}_{1} \in \mathbb{R}^{2p \times p}$, $W^{Joint}_{2} \in \mathbb{R}^{p \times p}$, $\hat{W} \in \mathbb{R}^{p}$ is the final weight for each path.

\subsection{Optimization}
\label{sec:objectives}

DTNet can be used for various V\&L downstream applications. For image captioning, we first pre-train our model with Cross-Entropy (CE) loss, which is formulated as:
\begin{equation}
    L_{CE}=-\sum_{t=1}^T \log\Big(p_\theta\left(y^*_t|y^*_{1:t-1}\right)\Big),
\label{eq:ce}
\end{equation}
where $y^*_{1:T}$ is the ground-truth caption with $T$ words, $\theta$ represents the parameter of our model.

Then, the model is optimized following Self-Critical Sequence Training (SCST) \cite{rennie2017self} according to the sum of CIDEr \cite{vedantam2015cider} and BLEU-4 \cite{papineni2002bleu}:
\begin{equation}
\nabla_{\theta} L_{RL}(\theta) = -\frac{1}{k} \sum_{i=1}^{k} \left(r\left(y_{1:T}^{i}\right) - b \right) \nabla_{\theta} \log p_{\theta}\left(y_{1:T}^{i}\right),
\label{eq:scst}
\end{equation}
where $k$ is the beam size, $r(\cdot)$ represents the reward, and $b = \Big(\sum_i r(y^i_{1:T})\Big)/k$ denotes the reward baseline.

\section{Experiment}

\subsection{Datasets and Experimental Settings}

We evaluate our proposed method on the popular image captioning benchmark MS-COCO \cite{lin2014microsoft},  containing more than 120,000 images. Concretely, it includes 82,783 training images, 40,504 validation images, and 40,775 testing images, each of which is annotated with 5 captions. For offline evaluation, we adopt the Karpathy split \cite{karpathy2015deep} where 5,000 images are used for validation, 5000 images for testing, and the rest images for training. For online evaluation, we upload the generated captions of the COCO official testing set to the online server.

The visual features are extracted from the Faster R-CNN \cite{ren2015faster} provided by Jiang \emph{et al.} \cite{jiang2020defense}. To reduce the computational overhead of Self-Attention, we average-pool features to 7×7 grid size following Luo \emph{et al.} \cite{luo2021dual}.

For fair comparisons, we use similar experimental settings to classic methods like \cite{luo2021dual,zhang2021rstnet,cornia2020meshed}. Concretely, $d_{model}$ is 512, the number of heads is 8, the expansion ratio of FFN is 4, the beam size of 5, the optimizer is Adam  \cite{kingma2014adam} and the number of layers in encoder and decoder is 3. Note that we do not use any extra data preprocessing, except simple augmentations (\emph{e.g.,} RandomCrop, RandomRotation). In the CE training stage, the batch size is 50, and the learning rate is linearly increased to $1 \times 10^{\text{-}4}$ during the first 4 epochs. Afterwards, we set it to $2 \times 10^{\text{-}5}$, $4 \times 10^{\text{-}6}$ at 10-th and 12-th epoch. After 18 epochs of CE pre-training, we choose the checkpoint achieving the best  CIDEr score for SCST optimization with the batch size of 100 and learning rate of $5 \times 10^{\text{-}6}$. The learning rate will be set to $2.5 \times 10^{\text{-}6}$, $5 \times 10^{\text{-}7}$, $2.5 \times 10^{\text{-}7}$, $5 \times 10^{\text{-}8}$ at the 35-th, 40-th, 45-th, 50-th epoch, and the SCST training will last 42 epochs.

Following the standard evaluation protocol, we utilized popular captioning metrics to evaluate our model, including BLEU-N \cite{papineni2002bleu}, METEOR \cite{banerjee2005meteor}, ROUGE \cite{lin2004rouge}, CIDEr \cite{vedantam2015cider} and SPICE \cite{anderson2016spice}.

\begin{table}[t]
\caption{
    Ablations on spatial modeling cells. All values are reported as percentage  (\%).  B-N,  M,  R,  C, and S are short for BLEU-N,  METEOR,  ROUGE-L,  CIDEr-D, and SPICE scores. GMC, LMC and AMC are short for Global Modeling Cell, Local Modeling Cell and Axial Modeling Cell, respectively.
}
\resizebox{1.00\columnwidth}{!}{
\begin{tabular}{ccc|cccccc}
\hline
GMC & LMC & AMC & B1            & B4            & M             & R             & C              & S             \\
\hline
  \textbf{$\times$}  &  \textbf{$\times$}   &  \textbf{$\times$}   & 80.8          & 39.5          & 29.3          & 58.8          & 132.5          & 22.7          \\
\textbf{$\surd$}   &   \textbf{$\times$}  &  \textbf{$\times$}   & 81.4          & 39.9          & 29.4          & 59.1          & 133.3          & 23.0          \\
  \textbf{$\times$}  & \textbf{$\surd$}   &  \textbf{$\times$}   & 81.1          & 39.6          & 29.4          & 59.0          & 132.7          & 22.9          \\
  \textbf{$\times$}  &  \textbf{$\times$}   & \textbf{$\surd$}
   & 81.1          & 39.7          & 29.4          & 58.9          & 133.4          & 22.9          \\ 

\color{black}{\textbf{$\surd$}}  &  \color{black}{\textbf{$\surd$}}  & \color{black}{\textbf{$\times$}}
& \color{black}{81.4}          & \textbf{\color{black}{40.0}}          & \textbf{\color{black}{29.5}}          & \color{black}{59.1}          & \color{black}{134.0}          & \textbf{\color{black}{23.1}}          \\ 
\color{black}{\textbf{$\times$}}  &  \color{black}{\textbf{$\surd$}}   & \color{black}{\textbf{$\surd$}}
& \color{black}{81.3}         & \color{black}{39.9}          & \color{black}{29.4}         & \color{black}{59.1}          & \color{black}{134.2}          &\textbf{\color{black}{23.1}}          \\ 
\color{black}{\textbf{$\surd$}}  &  \color{black}{\textbf{$\times$}}   & \color{black}{\textbf{$\surd$}}
&    \color{black}{81.4}      & \color{black}{39.9}          & \color{black}{29.4}         & \color{black}{59.1}          & \color{black}{134.3}          & \color{black}{23.0}          \\

\hline
\textbf{$\surd$}   & \textbf{$\surd$}   & \textbf{$\surd$}   & \textbf{81.5} & \textbf{40.0} & \textbf{29.5} & \textbf{59.2} & \textbf{134.9} & \textbf{23.1} \\
\hline
\end{tabular}
}
\vspace{-0.0cm}

\vspace{-0.0cm}
\label{tab:spatial}
\end{table}

\begin{table}[t]
\caption{Ablation studies on channel modeling cells. B-1, B-4, M, R, C, and S are short for BLEU-1, BLEU-4, METEOR, ROUGE, CIDEr, SPICE scores, respectively. CAC and CPC are short for Channel Attention Cell and Channel Projection Cell.}
\resizebox{1.00\columnwidth}{!}{
\begin{tabular}{cc|cccccc}
\hline
CAC & CPC & B1            & B4            & M             & R             & C              & S             \\ \hline
  \textbf{$\times$}  &  \textbf{$\times$}   & 80.9          & 39.2          & 29.0          & 58.7          & 132.1          & 22.6          \\
\textbf{$\surd$}   &   \textbf{$\times$}  & 81.0          & 39.3          & 29.1          & 58.9          & 132.9          & 22.7          \\
  \textbf{$\times$}  & \textbf{$\surd$}   & 81.0          & 39.5          & 29.4          & 59.0          & 133.1          & 23.0          \\ \hline
\textbf{$\surd$}   & \textbf{$\surd$}   & \textbf{81.5} & \textbf{40.0} & \textbf{29.5} & \textbf{59.2} & \textbf{134.9} & \textbf{23.1}\\
\hline
\end{tabular}
}
\vspace{-0.0cm}
\vspace{-0.0cm}
\label{tab:channel}
\end{table}

\begin{table}[t]
\caption{
    Ablations on various arrangements of dynamic spatial and channel blocks. `\emph{S}' and `\emph{C}' are short for Spatial and Channel. `\emph{\&}' and `\emph{+}' represent parallel and sequential connections, respectively. B-1, B-4, M, R, C, and S are short for BLEU-1, BLEU-4, METEOR, ROUGE, CIDEr, SPICE scores, respectively
}
\resizebox{1.00\columnwidth}{!}{
\begin{tabular}{c|cccccc}
\hline
Arrangements & B1   & B4   & M    & R    & C     & S    \\ \hline
\emph{S \& C}  & 81.1 & \textbf{40.0} & 29.3 & 59.1 & 133.6 & 22.7 \\
\emph{C + S}   & 81.3 & 39.9 & 29.3 & 58.9 & 133.8 & 23.0 \\
\emph{S + C}   & \textbf{81.5} & \textbf{40.0} & \textbf{29.5} & \textbf{59.2} & \textbf{134.9} & \textbf{23.1} \\ \hline
\end{tabular}
}
\vspace{-0.0cm}

\vspace{-0.0cm}
\label{tab:arrangement}
\end{table}

\begin{table}[t]
\caption{Ablation studies on various routers.  B-1, B-4, M, R, C, and S are short for BLEU-1, BLEU-4, METEOR, ROUGE, CIDEr, SPICE scores, respectively}
\resizebox{1.00\columnwidth}{!}{
\begin{tabular}{c|cccccc}
\hline
Router & B1   & B4   & M    & R    & C     & S    \\ \hline
Static Summation & 81.0 &	39.4 &	29.3 &	59.0 &	133.0 &	22.7 \\ 
Spatial-based & 81.1          & 39.5          & 29.4          & 59.0          & 133.1          & 23.0          \\
Channel-based & 81.1          & 39.8          & \textbf{29.5}          & 59.0          & 133.6          & 23.1          \\ \hline
SCJR          & \textbf{81.5} & \textbf{40.0} & \textbf{29.5} & \textbf{59.2} & \textbf{134.9} & \textbf{23.1} \\ \hline
\end{tabular}
}
\vspace{-0.0cm}

\vspace{-0.0cm}
\label{tab:router}
\end{table}

\begin{table}[t]
\caption{Ablation studies on the grouping operation for cells.  B-1, B-4, M, R, C, and S are short for BLEU-1, BLEU-4, METEOR, ROUGE, CIDEr, SPICE scores, respectively.}
\resizebox{1.00\columnwidth}{!}{
\begin{tabular}{c|cccccc}
\hline
Grouping & B1            & B4            & M             & R             & C              & S             \\ \hline
\textbf{$\times$}      & 81.2          & 39.7          & 29.4          & 59.0          & 133.5          & 23.0          \\
\textbf{$\surd$}      & \textbf{81.5} & \textbf{40.0} & \textbf{29.5} & \textbf{59.2} & \textbf{134.9} & \textbf{23.1} \\ \hline
\end{tabular}
}
\vspace{-0.0cm}

\vspace{-0.0cm}
\label{tab:grouping}
\end{table}

\begin{table}[t]
\caption{{\color{black}{Performance comparison with different grouping combinations.  B-1, B-4, M, R, C, and S are short for BLEU-1, BLEU-4, METEOR, ROUGE, CIDEr, and SPICE scores, respectively.}}}
\resizebox{1.00\columnwidth}{!}{
\begin{tabular}{l|l|cccccc}
\hline
Group1        & Group2   & B-1           & B-4           & M             & R             & C              & S             \\ \hline
CAC, LMC, AMC & CPC, GMC & 81.0          & 39.8          & 29.3          & \textbf{59.2} & 133.6          & 23.0          \\
CAC, GMC, AMC & CPC, LMC & 81.4          & \textbf{40.0} & \textbf{29.5} & 59.1          & 133.9          & 23.0          \\
CAC, GMC, LMC & CPC, AMC & 81.3          & 39.6          & \textbf{29.5} & 59.1          & 134.2          & \textbf{23.1} \\
CPC, LMC, AMC & CAC, GMC & 81.4          & 39.9          & 29.4          & 59.2          & 133.6          & 22.9          \\
CPC, GMC, AMC & CAC, LMC & \textbf{81.5} & \textbf{40.0} & 29.3          & 59.1          & 134.0          & 23.0          \\
CPC, GMC, LMC & CAC, AMC & 81.4          & 39.8          & 29.4          & 59.1          & 133.9          & 23.0          \\
GMC, LMC, AMC & CAC, CPC & \textbf{81.5} & \textbf{40.0} & \textbf{29.5} & \textbf{59.2} & \textbf{134.9} & \textbf{23.1} \\ \hline
\end{tabular}
}
\label{tab:group_option}
\end{table}

\subsection{Ablation Analysis}

\subsubsection{Ablation on Spatial Modeling Cells}
To gain insights into three spatial modeling cells, we conduct detailed ablation studies. As shown in Tab. \ref{tab:spatial}, we observe that whichever cell is equipped, the performance will be significantly improved, which proves the effectiveness of our proposed cells. Moreover, compared with LMC and AMC, GMC achieves better performance, which indicates that  global modeling plays a more important role than the local and axial one. 
{\color{black}{
Furthermore, we can observe that the simultaneous utilization of two spatial modeling cells enhances performance in comparison to exclusively relying on one type. For example, when both the GMC and AMC are engaged jointly, we note an appreciable increase in CIDEr scores; as measured, there is a 1.0 CIDEr and 0.9 CIDEr increment over the utilization of solely the GMC or AMC, respectively. Additionally, we find that uniting all three spatial modeling cells - the GMC, LMC, and AMC - garners even more significant gains. This phenomenon can be ascribed to the synergistic effect of global, local, and axial modeling operating in the spatial domain. Together, these different modeling techniques collectively enhance the understanding of visual semantics within an image. Consequently, this coordination aids in generating more accurate and fluid image captions.
}}
{\color{black}{
 Critically, an improved score of 2.4 CIDEr (i.e., from 132.5 to 134.9) is evident in the experiments with our three proposed spatial modeling cells.  This indicates that these cells provide an effective mechanism for spatial information modeling.
}}

\subsubsection{Ablation on Channel Modeling Cells}
To explore the impact of channel modeling cells, we also conduct ablation studies incrementally. As reported in Tab. \ref{tab:channel}, we can observe that equipping channel modeling cells also contributes to better performance. Specifically, CAC and CPC help the captioning model achieve 0.8\% and 1.0\% improvement on the CIDEr score, so both attention-based cell and projection-based cell can improve the semantic modeling ability of the model and the accuracy of the generated captions. Besides, equipping both channel modeling cells can further push the performance, \emph{i.e.,} 2.8\% improvement on the CIDEr score. Although both CAC and CPC are the modeling modules in the channel domain, because their modeling principles are different (\emph{i.e.,} attention-based method and projection-based method), they can promote each other to achieve higher performance.
{\color{black}{
 Importantly, Tab.~\ref{tab:channel} reveals an enhancement of 2.8 in the CIDEr score (from 132.1 to 134.9) due to our two proposed channel modeling cells, thereby demonstrating their efficacy at channel information modeling.
}}

\begin{table}[t]
\caption{Ablation studies on various routing types.  B-1, B-4, M, R, C, and S are short for BLEU-1, BLEU-4, METEOR, ROUGE, CIDEr, SPICE scores, respectively}
\resizebox{1.00\columnwidth}{!}{
\begin{tabular}{c|cccccc}
\hline
 Type & B1            & B4            & M             & R             & C              & S             \\ \hline
Static & 81.0 &	39.4 &	29.3 &	59.0 &	133.0 &	22.7 \\
Hard   Routing      & 81.4          & \textbf{40.1}          & 29.3          & 59.1          & 133.3          & 22.8          \\
Soft    Routing     & \textbf{81.5} & 40.0 & \textbf{29.5} & \textbf{59.2} & \textbf{134.9} & \textbf{23.1} \\ \hline
\end{tabular}
}
\vspace{-0.0cm}

\vspace{-0.0cm}
\label{tab:routing}
\end{table}

\subsubsection{Effect of Different Cell Arrangements}
To explore the effect of different arrangements of modeling cells, we compare three ways for arranging spatial and channel modeling cells:  parallel channel-spatial (\emph{S\&C}),  sequential channel-spatial (\emph{C+S}) and sequential spatial-channel (\emph{S+C}). Tab. \ref{tab:arrangement} summarizes the results of different arrangement methods. By analyzing the experimental results, we can find that \emph{S+C} performs consistently better than \emph{S\&C} and \emph{C+S}.

\begin{table}[t]
\begin{small}
\caption{Comparisons with SOTAs on the Karpathy test split.  B-1, B-4, M, R, C, and S are short for BLEU-1, BLEU-4, METEOR, ROUGE, CIDEr, SPICE scores, respectively}
\resizebox{1.00\columnwidth}{!}{
	\begin{tabular}{l|c c c c c c}
		\hline
		    Model        &B-1  & B-4& M & R & C & S \\ 
		\hline

    \multicolumn{7}{l}{\cellcolor[HTML]{C0C0C0}\textbf{\color{black}{Large-scale Vision Language Pre-Training Models}}} \\ \hline

\color{lightgray}{CLIP-ViL~\cite{shen2021much}} & \color{lightgray}{-}& \color{lightgray}{40.2}& \color{lightgray}{29.7}& \color{lightgray}{-}& \color{lightgray}{134.2}& \color{lightgray}{23.8}  \\

 \color{lightgray}{BLIP~\cite{li2022blip}} & \color{lightgray}{-}& \color{lightgray}{40.4}& \color{lightgray}{-}& \color{lightgray}{-}& \color{lightgray}{136.7}& \color{lightgray}{-}  \\

\color{lightgray}{VinVL~\cite{zhang2021vinvl}} & \color{lightgray}{-}& \color{lightgray}{41.0}& \color{lightgray}{31.1}& \color{lightgray}{-}& \color{lightgray}{140.9}& \color{lightgray}{25.4}  \\

\color{lightgray}{OSCAR~\cite{li2020oscar}} & \color{lightgray}{-}& \color{lightgray}{41.7}& \color{lightgray}{30.6}& \color{lightgray}{-}& \color{lightgray}{140.0}& \color{lightgray}{24.5}  \\


\color{lightgray}{LEMON~\cite{hu2022scaling}} & \color{lightgray}{-}& \color{lightgray}{42.3}& \color{lightgray}{31.2}& \color{lightgray}{-}& \color{lightgray}{144.3}& \color{lightgray}{25.3}  \\

\color{lightgray}{OFA~\cite{wang2022ofa}} & \color{lightgray}{-}& \color{lightgray}{43.5}& \color{lightgray}{31.9}& \color{lightgray}{-}& \color{lightgray}{149.6}& \color{lightgray}{26.1}  \\




\hline
\multicolumn{7}{l}{\cellcolor[HTML]{C0C0C0}\textbf{\color{black}{Image Captioning Models without Pretraining}}} \\ \hline
  
        SCST  \cite{rennie2017self}     &   -  & 34.2 & 26.7 & 55.7 & 114.0&  -   \\ 
         Up-Down \cite{anderson2018bottom}   & 79.8 & 36.3 & 27.7 & 56.9 & 120.1& 21.4 \\ 
         RFNet \cite{jiang2018recurrent}     & 79.1 & 36.5 & 27.7 & 57.3 & 121.9& 21.2 \\ 
         GCN-LSTM  \cite{yao2018exploring}    & 80.5 & 38.2 & 28.5 & 58.3 & 127.6& 22.0 \\ 
         SGAE  \cite{yang2019auto}       & 80.8 & 38.4 & 28.4 & 58.6 & 127.8& 22.1 \\
         AoANet   \cite{huang2019attention}     & 80.2 & 38.9 & 29.2 & 58.8 & 129.8& 22.4 \\ 
         ORT    \cite{herdade2019image}    & 80.5 & 38.6 & 28.7 & 58.4 & 128.3& 22.6 \\ 
         Transformer  \cite{vaswani2017attention} & 81.0          & 38.9          & 29.0          & 58.4          & 131.3          & 22.6  \\ 
         $M^2$Transformer  \cite{cornia2020meshed}  & 80.8 & 39.1 & 29.2 & 58.6 & 131.2& 22.6 \\ 
         XTransformer \cite{pan2020x}  & 80.9 & 39.7 & {29.5} & 59.1 & 132.8& \textbf{23.4} \\ 
          
          DLCT \cite{luo2021dual} & 81.4& 39.8 &{29.5}&59.1& 133.8 & 23.0 \\
          RSTNet \cite{zhang2021rstnet}   & 81.1 & 39.3& 29.4& 58.8& 133.3 &23.0 \\ 

\color{black}{CMAL~\cite{guo2020non}} & \color{black}{80.3}& \color{black}{37.3}& \color{black}{28.1}& \color{black}{58.0}& \color{black}{124.0}& \color{black}{21.8} \\

\color{black}{SATIC~\cite{zhou2021semi}} & \color{black}{80.6}& \color{black}{37.9}& \color{black}{28.6}& \color{black}{-}& \color{black}{127.2}& \color{black}{22.3} \\

\color{black}{TCIC~\cite{fan2021tcic}} & \color{black}{80.9}& \color{black}{39.7}& \color{black}{29.2}& \color{black}{58.6}& \color{black}{132.9}& \color{black}{22.4} \\

\color{black}{DeeCap~\cite{fei2022deecap}} & \color{black}{80.1}& \color{black}{38.7}& \color{black}{29.1}& \color{black}{58.1}& \color{black}{129.0}& \color{black}{22.5} \\

\color{black}{TRRAR~\cite{wang2022text}} & \color{black}{80.1}& \color{black}{38.7}& \color{black}{28.8}& \color{black}{58.8}& \color{black}{128.0}& \color{black}{22.6} \\

\color{black}{$S^2$Transformer~\cite{zeng2022s2}} & \color{black}{81.1}& \color{black}{39.6}& \color{black}{29.6}& \color{black}{59.1}& \color{black}{133.5}& \color{black}{23.2} \\

\color{black}{UAIC~\cite{fei2023uncertainty}} & \color{black}{80.9}& \color{black}{38.8}& \color{black}{29.2}& \color{black}{58.7}& \color{black}{131.7}& \color{black}{22.8} \\

\color{black}{SCD-Net~\cite{luo2023semantic}} & \color{black}{81.3}& \color{black}{39.4}& \color{black}{29.2}& \color{black}{59.1}& \color{black}{131.6}& \color{black}{23.0} \\

\color{black}{MAN~\cite{jing2023memory}} & \color{black}{81.0}& \color{black}{39.4}& \color{black}{29.5}& \color{black}{59.0}& \color{black}{133.3}& \color{black}{23.1} \\


          \hline 
          
         DTNet (Ours) &  \textbf{81.5}          & \textbf{40.0} & \textbf{29.5} & \textbf{59.2} & \textbf{134.9} & 23.1  \\
        \hline
        \end{tabular}
}
    \end{small}
    \vspace{-0.0cm}
	
	\vspace{-0.0cm}
    \label{tab:offline_sota}
\end{table}

\begin{table*}[t]
    \small
	\centering
	\caption{Leaderboard of the published state-of-the-art image captioning models on the COCO online testing server. ${\dag}$ represents adopting both grid and region visual features.}
	\resizebox{2.00\columnwidth}{!}{
	\begin{tabular}{l|cccccccccccccc}
		\hline
		\multirow{2}{*}{Model}    & \multicolumn{2}{c}{BLEU-1}   & \multicolumn{2}{c}{BLEU-2}   & \multicolumn{2}{c}{BLEU-3}   & \multicolumn{2}{c}{BLEU-4}   & \multicolumn{2}{c}{METEOR}   & \multicolumn{2}{c}{ROUGE-L}  & \multicolumn{2}{c}{CIDEr-D}    \\ 
		    & c5            & c40           & c5            & c40           & c5            & c40           & c5            & c40           & c5            & c40           & c5            & c40           & c5             & c40            \\ \hline
		SCST   \cite{rennie2017self}      & 78.1          & 93.7          & 61.9          & 86.0          & 47.0          & 75.9          & 35.2          & 64.5          & 27.0          & 35.5          & 56.3          & 70.7          & 114.7          & 116.0          \\ 
		LSTM-A  \cite{yao2017boosting}  & 78.7          & 93.7          & 62.7          & 86.7          & 47.6          & 76.5          & 35.6          & 65.2          & 27.0          & 35.4          & 56.4          & 70.5          & 116.0          & 118.0          \\ 
		Up-Down  \cite{anderson2018bottom}   & 80.2          & 95.2          & 64.1          & 88.8          & 49.1          & 79.4          & 36.9          & 68.5          & 27.6          & 36.7          & 57.1          & 72.4          & 117.9          & 120.5          \\ 
		RF-Net \cite{jiang2018recurrent}   & 80.4          & 95.0          & 64.9          & 89.3          & 50.1          & 80.1          & 38.0          & 69.2          & 28.2          & 37.2          & 58.2          & 73.1          & 122.9          & 125.1          \\ 
		GCN-LSTM  \cite{yao2018exploring}  & 80.8             & 95.2             & 65.5          & 89.3          & 50.8          & 80.3          & 38.7          & 69.7          & 28.5          & 37.6          & 58.5          & 73.4          & 125.3          & 126.5          \\ 
		SGAE  \cite{yang2019auto}  & 81.0          & 95.3          & 65.6          & 89.5          & 50.7          & 80.4          & 38.5          & 69.7          & 28.2          & 37.2          & 58.6          & 73.6          & 123.8          & 126.5          \\ 
		AoANet  \cite{huang2019attention}  & 81.0          & 95.0          & 65.8          & 89.6          & 51.4          & 81.3          & 39.4          & 71.2          & 29.1          & 38.5          & 58.9          & 74.5 & 126.9          & 129.6          \\ 
		ETA   \cite{li2019entangled}    & 81.2          & 95.0          & 65.5          & 89.0          & 50.9          & 80.4          & 38.9          & 70.2          & 28.6          & 38.0          & 58.6          & 73.9          & 122.1          & 124.4          \\
		$M^2$Transformer  \cite{cornia2020meshed}    & 81.6         & 96.0          & 66.4          & 90.8          & 51.8          & 82.7          & 39.7          & 72.8          & 29.4          & 39.0          & 59.2         & 74.8          & 129.3          & 132.1          \\
		XTransformer  \cite{pan2020x}  (ResNet-101)    & 81.3          & 95.4          & 66.3          & 90.0          & 51.9          & 81.7          & 39.9          & 71.8          & 29.5          & 39.0          & 59.3          & 74.9          & 129.3          & 131.4          \\ 
		XTransformer  \cite{pan2020x} (SENet-154)   &81.9  &95.7  &66.9  &90.5  &52.4  &82.5  &40.3  &72.4  &29.6  &39.2  &59.5  &75.0  &131.1  &133.5\\

		RSTNet \cite{zhang2021rstnet}(ResNeXt101)  &81.7  &96.2  &66.5  &90.9 & 51.8 & 82.7 & 39.7 & 72.5 & 29.3 & 38.7 & 59.2 & 74.2 & 130.1 & 132.4 \\
		RSTNet \cite{zhang2021rstnet}(ResNeXt152) & 82.1 & 96.4 & 67.0 & 91.3 & 52.2 & 83.0 & 40.0 & 73.1 & 29.6 & 39.1 & 59.5 & 74.6 & 131.9 & 134.0\\
		
        DLCT $^{\dag}$ \cite{luo2021dual} (ResNeXt101)  &82.0 & 96.2 & 66.9 & 91.0 & 52.3 & 83.0 & 40.2 & 73.2 & 29.5 & 39.1 & 59.4 & 74.8 & 131.0 & 133.4\\
        DLCT $^{\dag}$ \cite{luo2021dual} (ResNeXt152) & 82.4 & {96.6} & 67.4 & 91.7 & 52.8 & 83.8 & 40.6 & 74.0 & {29.8} & \textbf{39.6} & {59.8} & \textbf{75.3} & 133.3 & 135.4\\

\color{black}{DeeCap~\cite{fei2022deecap}} & \color{black}{80.5}& \color{black}{95.1}& \color{black}{65.2}& \color{black}{ 89.1}& \color{black}{50.3}& \color{black}{80.0}& \color{black}{38.1}& \color{black}{69.5}& \color{black}{28.0}& \color{black}{37.0}& \color{black}{58.4}& \color{black}{73.5}& \color{black}{121.4} & \color{black}{124.4} \\

\color{black}{TRRAR~\cite{wang2022text}} & \color{black}{80.2} & \color{black}{94.7} & \color{black}{64.9} & \color{black}{88.9} & \color{black}{50.4} & \color{black}{80.3} & \color{black}{38.5} & \color{black}{70.0} & \color{black}{29.0} & \color{black}{38.4} & \color{black}{58.7} & \color{black}{74.2} & \color{black}{125.1} & \color{black}{127.6}  \\

\color{black}{$A^2$Transformer} & \color{black}{82.2} & \color{black}{96.4} & \color{black}{67.0} & \color{black}{91.5} & \color{black}{52.4} & \color{black}{83.6} & \color{black}{40.2} & \color{black}{73.8} & \color{black}{29.7} & \color{black}{39.3} & \color{black}{59.5} & \color{black}{75.0} & \color{black}{132.4} & \color{black}{134.7}  \\

\color{black}{SCD-Net~\cite{luo2023semantic}} & \color{black}{80.2} & \color{black}{95.1} & \color{black}{67.0} & \color{black}{89.3} & \color{black}{50.1} & \color{black}{80.1} & \color{black}{38.1} & \color{black}{69.4} & \color{black}{29.0} & \color{black}{38.2} & \color{black}{58.5} & \color{black}{73.5} & \color{black}{126.2} & \color{black}{129.2}  \\

\color{black}{UAIC~\cite{fei2023uncertainty}} & \color{black}{81.9} & \color{black}{96.3} & \color{black}{66.5} & \color{black}{91.1} & \color{black}{51.8} & \color{black}{83.0} & \color{black}{39.6} & \color{black}{72.9} & \color{black}{29.2} & \color{black}{38.9} & \color{black}{59.2} & \color{black}{74.7} & \color{black}{129.0} & \color{black}{132.8}  \\


		 \hline
		DTNet  (ResNeXt-101)  & 82.1 & 96.2 & 67.0 & 91.2 & 52.5 & 83.3 & 40.5 & 73.5 & 29.5 & 39.1 & 59.5 & 74.8 & 131.6 & 133.9 \\
		
		DTNet  (ResNeXt-152)     & \textbf{82.5} & \textbf{96.6} & \textbf{67.6} & \textbf{91.9} & \textbf{53.2} & \textbf{84.1} & \textbf{41.0} & \textbf{74.3} & \textbf{29.8} & 39.5 & \textbf{59.8} & 75.2 & \textbf{133.9} & \textbf{136.1} \\
		\hline
	\end{tabular}
}
    \vspace{-0.0cm}

	\vspace{-0.0cm}
	\label{tab:online_sota}
\end{table*}

\begin{table}[t]
	\caption{ Comparisons with SOTA methods on the Karpathy test split using the same ResNeXt-101 grid feature.  B-1, B-4, M, R, C, and S are short for BLEU-1, BLEU-4, METEOR, ROUGE, CIDEr, SPICE scores, respectively}
	\resizebox{1.00\columnwidth}{!}{
	\begin{tabular}{l|c c c c c c}
		\hline
	    Model        &B-1  & B-4& M & R & C & S \\ 
	    \hline
		Up-Down \cite{anderson2018bottom}      & 75.0 & 37.3 &28.1  & 57.9 & 123.8& 21.6 \\ 
         AoANet   \cite{huang2019attention}     & 80.8 & 39.1 &29.1  & 59.1 & 130.3& 22.7 \\ 
 		 Transformer \cite{vaswani2017attention} & 81.0 & 38.9 &29.0  & 58.4 & 131.3& 22.6 \\ 
 		 $M^2$Transformer  \cite{cornia2020meshed}  & 80.8 & 38.9 &29.1  & 58.5 & 131.8& 22.7 \\ 
        XTransformer  \cite{pan2020x} & 81.0 & 39.7 &29.4  & 58.9 & 132.5& \textbf{23.1} \\
        DLCT  \cite{luo2021dual}  & 81.1 & 39.3 & 29.4 & 58.9 & 132.5 & 22.9 \\   
        RSTNet  \cite{zhang2021rstnet}  & 81.1 & 39.3& 29.4& 58.8& 133.3 &23.0 \\ \hline  
         DTNet (Ours)&  \textbf{81.5}          & \textbf{40.0} & \textbf{29.5} & \textbf{59.2} & \textbf{134.9} & \textbf{23.1}         \\
		
		\hline
	\end{tabular}
    	}
	\vspace{-0.0cm}

	\label{tab:grid}
	\vspace{-0.0cm}
\end{table}

\subsubsection{Effect of Different Routers}
Different from previous works, where routers are based on Squeeze-and-Excitation \cite{hu2018squeeze}, our proposed SCJR executes the path customization according to both channel and spatial information of input samples. To verify its efficacy, we conduct extensive ablation experiments by decoupling spatial and channel branches of SCJR. Besides, we also report the performance of ``static  router'', which directly sums the outputs of all cells. As reported in Tab. \ref{tab:router}, we observe that our proposed SCJR performs better than the spatial-based and channel-based routers by a notable margin, which confirms the importance of joint modeling in both spatial and channel domains. Particularly, SCJR outperforms the spatial-based and channel-based router by \textbf{\emph{1.8\%}} and \textbf{\emph{1.3\%}} on the CIDEr score. Note that all dynamic routers perform better than the ``static router'', showing that dynamic routing is critical for pushing performance in image captioning.

\subsubsection{Effect of the Grouping Operation of Cells}
To explore the impact of the grouping operation for cells, we also conduct experiments by placing all spatial and channel modeling cells in the same routing space. As shown in Tab. \ref{tab:grouping}, we could observe that performance drops significantly (\emph{i.e.,} \textbf{\emph{1.4\%}} on the CIDEr score) without grouping operation. The reason may be that spatial and channel cells are complementary, and placing them in the same routing space will damage the routing efficiency. After they are grouped according to prior knowledge, the model no longer needs to decide whether to take the channel path or the spatial path, therefore reducing the optimization difficulty.

\subsubsection{Effect of Different Routing Types}
With \emph{Gumbel-Softmax Trick} \cite{jang2016categorical}, we also implement an end-to-end hard routing scheme, which achieves binary path selection in the encoder. As reported in Tab. \ref{tab:routing}, we could find that the hard routing model performs worse than the soft one, yet still outperforms the static one, which can be easily explained in terms of the number of sub-models. All samples go through the same path in the static model, so the number of sub-models in the static model is $1$. Similarly, because of binary path selection, the upper-bound number of sub-models in the hard routing model is $\Pi_{i=1}^L(N^i_sN^i_c)$, where $L$ is the number of encoder layers, $N_s^i$, $N_c^i$ are the number of spatial and channel modeling cells in the $i$-th layer. The soft routing model can assign different path weights based on input samples, so the upper-bound number of sub-models in the soft routing model is $+ \infty $.

\subsubsection{{\color{black}{Effect of Different Grouping Combinations}}}
{\color{black}{
To investigate the impact of different grouping combinations, we extensively examined a range of grouping configurations, which include diverse combinations of spatial modeling cells and channel modeling cells within the same routing space. Our empirical results, as illustrated in the first six rows of Tab.~\ref{tab:group_option}, consistently demonstrate that performance degradation occurs to differing extents when spatial and channel modeling cells are intermixed in the routing space. When we allocate these two classes of cells into separate routing spaces, the image captioning model is granted a focused attention on spatial and channel modeling, which we believe is the key to superior performance. This observation substantiates our earlier hypothesis that the functionalities of spatial modeling cells and channel modeling cells are mutually complementary, thereby signifying the crucial role of distinct groupings. With these empirically-grounded observations and subsequent analysis, we propose the segregation of the five basic cell types into two distinct groups, designed for spatial modeling and channel modeling, respectively.
}}

\subsection{General Performance Comparison}
\subsubsection{Offline Evaluation}
In Tab. \ref{tab:offline_sota}, we report the performance comparison between our proposed DTNet and previous SOTAs on the offline COCO Karpathy split.  For fair comparisons, we report the results of single models without using any ensemble technologies. As can be observed, our DTNet performs better than other models in terms of most metrics. Specifically, the CIDEr score of DTNet is \textbf{\emph{134.9 \%}}, outperforming all previous methods by a significant margin.

\begin{figure*}
\vspace{-0.0cm}
\centering 
  \includegraphics[width=1.80\columnwidth]{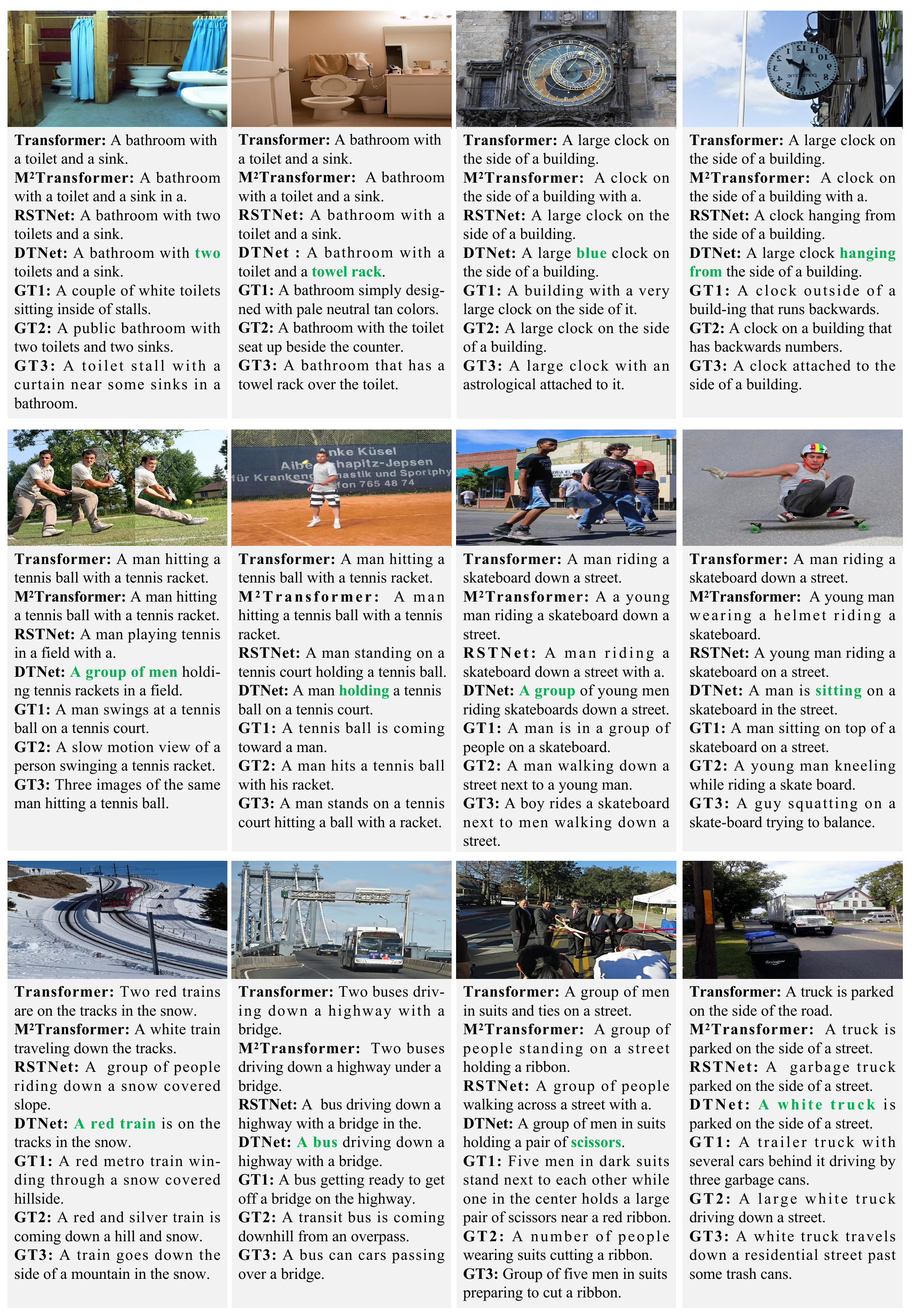}
  \vspace{-0.0cm}
  \caption{ 
     {\color{black}{Examples of captions generated by Transformer~\cite{vaswani2017attention}, $M^2$Transformer~\cite{cornia2020meshed}, RSTNet~\cite{zhang2021rstnet} and DTNet. ``GT'' is short for ``Ground Truth''.}}
  }
  \vspace{-0.0cm}
  \label{fig:fig5}
\end{figure*}

\begin{table*}[]
\caption{Performance comparisons of different captioning metrics for the Standard Transformer and our DTNet.  P-values come from two-tailed t-tests using paired samples. P-values in bold are significant at  0.05  significance level.}
\label{tab:p_metric}
\begin{center}

\begin{tabular}{l|cccccc}
\toprule
Model       & BLEU-1               & BLEU-4            & METEOR        & ROUGE & CIDEr & SPICE \\
\midrule
Transformer & 81.0            & 38.9          & 29.0            & 58.4  & 131.3 & 22.6  \\
DTNet        & 81.5& 40.0 & 29.5 & 59.2  & 134.9 & 23.1  \\
\hline
p-value & {$\mathbf{8.66 \times 10^{-3}}$}  & {$\mathbf{3.24 \times 10^{-5}}$} & {$\mathbf{2.12 \times 10^{-7}}$} & {$\mathbf{5.83 \times 10^{-6}}$} & {$\mathbf{5.50 \times 10^{-7}}$} & {$\mathbf{2.80 \times 10^{-5}}$}\\
\bottomrule
\end{tabular}
    
\end{center}
\end{table*}

\begin{table*}[]
\centering
\caption{ Subcategories of SPICE metrics for the Standard Transformer and our proposed DTNet. P-values are calculated by two-tailed t-tests using paired samples. Note that p-values in bold are significant at  0.05  significance level. }
\label{tab:spice}
    \begin{tabular}{l|cccccc}
    \toprule
    \multirow{2}{*}{Model} & \multicolumn{6}{c}{SPICE}                                  \\ \cline{2-7}
       & Relation & Cardinality & Attribute & Size & Color & Object \\ 
       \hline
      Transformer  &6.91           & 20.58          & 11.80          & 4.71  & 12.93          & 40.35             \\
      DTNet  &7.06           & 22.07          & 12.29          & 4.98  & 14.23          & 40.90             \\ \hline
       p-value & $2.83 \times 10^{-1}$ & $1.31 \times 10^{-1}$ & $\mathbf{1.48 \times 10^{-05}}$ & $6.38 \times 10^{-1}$ & {\color{black}{$1.40 \times 10^{-1}$}} & {$\mathbf{4.65 \times 10^{-4}}$} \\

    \bottomrule
    \end{tabular}
\end{table*}

\begin{table}
\centering
{\color{black}
\caption{Comparison with the state of the art on the Flickr8K dataset. All values are reported as percentage  (\%),  where B-N,  M,  R, and  C   are short for BLEU-N,  METEOR,  ROUGE-L, and  CIDEr scores. † indicates an ensemble model results.}
\label{tab:flickr8k}
\begin{tabular}{l|ccccc}
\toprule
Methods        & B1            & B4            & M             & R             & C             \\ \midrule
Deep VS \cite{karpathy2015deep}       & 57.9          & 16.0          & -             & -             & -             \\
Google NIC \cite{vinyals2015show}†   & 63.0          & -             & -             & -             & -             \\
Soft-Attention \cite{xu2015show} & 67.0          & 19.5          & 18.9          & -             & -             \\
Hard-Attention \cite{xu2015show} & 67.0          & 21.3          & 20.3          & -             & -             \\
emb-gLSTM  \cite{jia2015guiding}    & 64.7          & 21.2          & 20.6          & -             & -             \\
Log Bilinear  \cite{donahue2015long}  & 65.6          & 17.7          & 17.3          & -             & -             \\ \hline
DTNet          & \textbf{68.3} & \textbf{26.7} & \textbf{22.0} & \textbf{49.9} & \textbf{66.7} \\ \bottomrule
\end{tabular}
}
\end{table}

\subsubsection{Online Evaluation}
Tab. \ref{tab:online_sota} summarizes the performance of SOTAs and our approach on the online test server. Note that we adopt two common backbones (ResNeXt-101 and ResNeXt-152 \cite{xie2017aggregated}) and ensemble of four models following \cite{pan2020x,cornia2020meshed}. The results demonstrate that DTNet has achieved the best result so far on most evaluation metrics.  
{\color{black}{
The proposed DTNet demonstrates superiority over DLCT in the following aspects:
(1) Superior Training Efficiency with DTNet: In the realm of offline-acquired feature training, DTNet markedly outpaces DLCT. This notable distinction arises from DLCT's integration of both grid and region features, which inevitably compounds computational overhead. Conversely, DTNet, by strategically excluding region features, harnesses a computational velocity three times that of DLCT during the cross-entropy training phase. This optimized approach reduces inherent algorithmic complexity and propels training efficiency.
(2) Accelerated Inference in DTNet: DTNet's single-stage design ensures rapid inference, eclipsing DLCT. A significant source of DLCT's latency, as emphasized in \cite{jiang2020defense}, is its dependency on region feature extraction, particularly the time-intensive NMS operation, which consumes a staggering 98.3\% of the total inference duration. In contrast, DTNet, by adopting an end-to-end design and omitting region features, vastly enhances its inference throughput, setting a new benchmark for operational efficiency.
(3) DTNet's Simplified yet Efficient Architecture: Whereas DLCT necessitates intricate designs due to its high structural complexity, DTNet offers a refreshing simplicity with robust performance. The crux of DLCT's design challenge lies in formulating complex interaction mechanisms to capitalize on the complementarity between diverse features. DTNet, however, embarks on a novel trajectory by leveraging automated structure optimization. It mandates only a selection from a curated set of architectures, utilizing a sample-adaptive method to dynamically pinpoint the most efficacious structure.
(4) Finally, in terms of performance, despite DLCT utilizing multiple visual features, DTNet consistently outperforms DLCT across most evaluative metrics on the COCO online testing server. Notably, the majority of performance indicators unequivocally support the superiority of DTNet, with only a marginal difference observed in the METEOR-c40 metric. Therefore, DTNet effectively demonstrates an outstanding balance between efficiency and performance, clearly showcasing its superiority over DLCT.
}
}

\begin{figure*}
\vspace{-0.0cm}
\centering 
  \includegraphics[width=2.00\columnwidth]{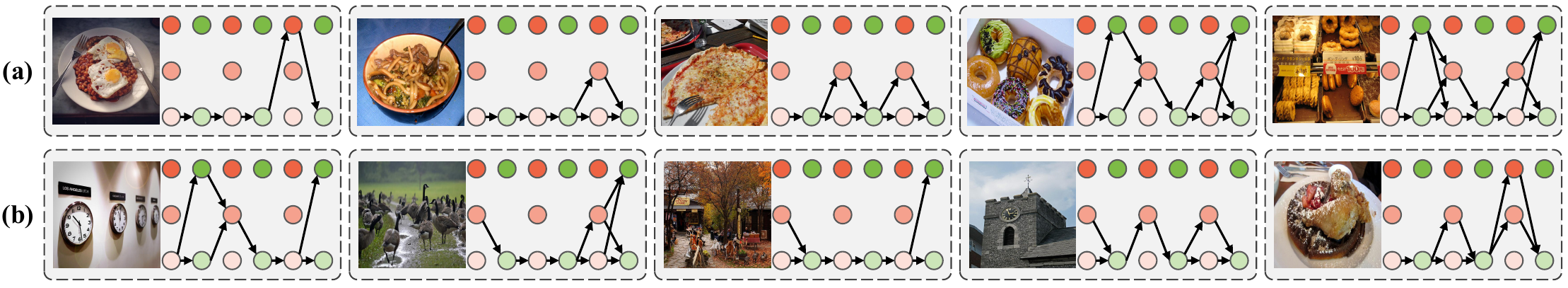}
  \vspace{-0.0cm}
  \caption{ 
     Path Visualization. (a) Examples in Fig. \ref{fig:fig7}. (b) Examples processed by the same number (\emph{i.e.,} 8) of cells.
  }
  \vspace{-0.0cm}
  \label{fig:fig8}
\end{figure*}

\subsubsection{Fair Comparisons with SOTA Methods}
To eliminate the interference caused by the adoption of different visual features, we also conduct extensive experiments on the same visual features to compare DTNet and previous SOTAs. As reported in Tab. \ref{tab:grid}, compared with other SOTAs, DTNet still shows significant performance superiority when using the same visual feature.

\subsection{Generalization On The Flickr Dataset}

We also perform extensive tests on the Flickr8K and Flickr30k datasets to validate the generalization of our proposed DTNet.

\begin{table}[t]
\centering
{\color{black}
\caption{Comparison with the state of the art on the Flickr30K dataset. All values are reported as percentage  (\%),  where B-N,  M,  R and  C   are short for BLEU-N,  METEOR,  ROUGE-L and  CIDEr scores. † indicates an ensemble model results.}
\label{tab:flickr30k}
\begin{tabular}{l|ccccc}
\toprule
Methods        & B1            & B4            & M             & R             & C             \\ \midrule
Deep VS   \cite{karpathy2015deep}      & 57.3          & 15.7          & -             & -             & -             \\
Google NIC \cite{vinyals2015show}†    & 66.3          & 18.3          & -             & -             & -             \\
m-RNN  \cite{mao2014deep}        & 60.0          & 19.0          & -             & -             & -             \\
Soft-Attention \cite{xu2015show} & 66.7          & 19.1          & 18.5          & -             & -             \\
Hard-Attention \cite{xu2015show} & 66.9          & 19.9          & 18.5          & -             & -             \\
emb-gLSTM   \cite{jia2015guiding}    & 64.6          & 20.6          & 17.9          & -             & -             \\
ATT \cite{you2016image}†           & 64.7          & 23.0            & 18.9          & -             & -             \\
Log Bilinear  \cite{donahue2015long}  & 60.0          & 17.1          & 16.9          & -             & -             \\ \hline
DTNet          & \textbf{70.1} & \textbf{25.7} & \textbf{20.9} & \textbf{48.1} & \textbf{59.0} \\ \bottomrule
\end{tabular}
}
\end{table}

\subsubsection{Performance Comparison on Flickr8K}

Flickr8K \cite{hodosh2013framing} is a collection of 8,000 images taken from Flickr. It includes five-sentence annotations for each image. The dataset provides a conventional separation of training, validation, and testing sets, which we use in the experiment. There are 6,000 training images, 1,000 validation images, and 1,000 testing images. Tab. \ref{tab:flickr8k} details the captioning performance of our proposed DTNet and previous approaches on the Flickr8K dataset. By analyzing the experimental results, we can observe that our proposed DTNet performs better than previous SOTAs. Notably, our proposed DTNet even outperforms some ensemble models (\emph{i.e.,} Google NIC \cite{vinyals2015show}).

\subsubsection{Performance Comparison on Flickr30K}

Flickr30K \cite{young2014image} is an extension to the Flickr8K collection. It also gives five-sentence annotations for each image. It has 158,915 captions from the public that describe 31,783 images. This dataset's annotations have similar grammar and style to Flickr8K. Following the previous research, we adopt 1,000 images for testing. Tab. \ref{tab:flickr30k} shows performance comparisons between our proposed DANet and prior SOTAs on the Flickr30K dataset. The outstanding performance of DTNet on Flickr30K again reveals the effectiveness and generalization of the dynamic network in the image captioning task.

\begin{figure}[t]
\centering 
  \includegraphics[width=1.00\columnwidth]{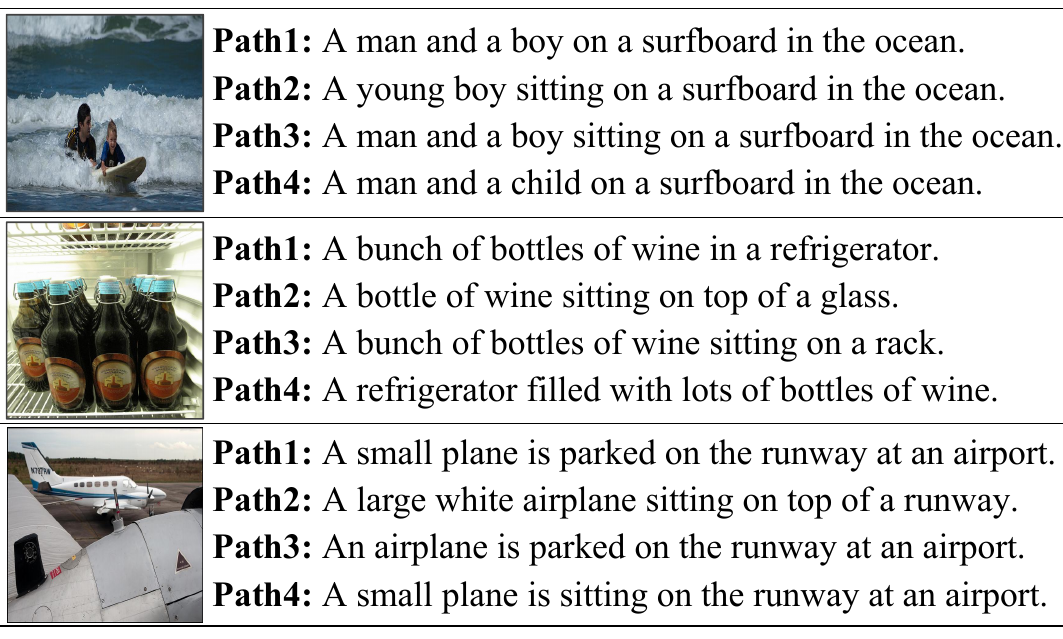}
  \vspace{-0.0cm}
  \caption{ 
     Captions generated by four random sampled paths from the proposed DTNet.
  }
  \vspace{-0.0cm}
  \label{fig:fig6}
\end{figure}

\begin{figure}
\vspace{-0.0cm}
\centering 
  \includegraphics[width=1.00\columnwidth]{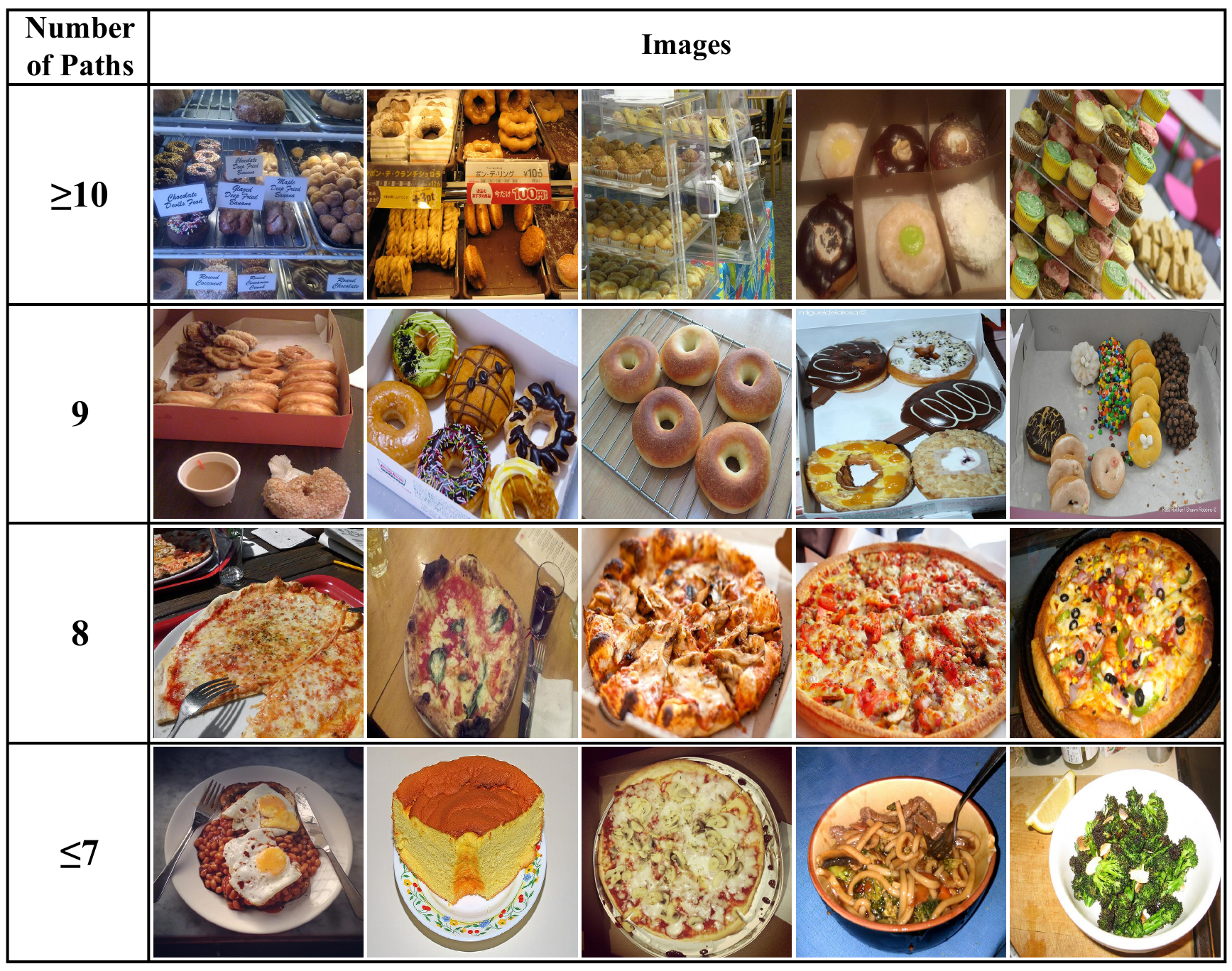}
  \vspace{-0.0cm}
  \caption{ 
     Images and the corresponding number of passed cells. Path visualization of some examples is shown in Fig. \ref{fig:fig8} (a).
  }
  \vspace{-0.0cm}
  \label{fig:fig7}
\end{figure}

\begin{table*}[t]
\centering
\caption{{\color{black}{Accuracies on the \emph{val} splits of VQA-v2 compared with SOTA approaches.}}}

\label{tab:vqa}
\setlength{\tabcolsep}{7mm}{
\begin{tabular}{l|cccc}
\toprule
 Model       & Overall  (\%) & Yes/No  (\%) & Number  (\%) & Other  (\%) \\ \midrule
 BUTD  \cite{teney2018tips}       & 63.84        & 81.40       & 43.81       & 55.78      \\
 MFB  \cite{yu2017multi}        & 65.35        & 83.23       & 45.31       & 57.05      \\
 MFH  \cite{yu2018beyond}        & 66.18        & 84.07       & 46.55       & 57.78      \\
BAN-4    \cite{kim2018bilinear}    & 65.86        & 83.53       & 46.36       & 57.56      \\
 BAN-8   \cite{kim2018bilinear}     & 66.00        & 83.61       & 47.04       & 57.62      \\
 MCAN  \cite{yu2019deep} & 67.17        & 84.82       & 49.31       & 58.48      \\

\color{black}{VL-T5~\cite{cho2021unifying}} & \color{black}{13.50}& \color{black}{-}& \color{black}{-}& \color{black}{-}\\

\color{black}{Frozen~\cite{tsimpoukelli2021multimodal}} & \color{black}{29.60}& \color{black}{-}& \color{black}{-}& \color{black}{-}\\

\color{black}{MetaLM~\cite{hao2022language}} & \color{black}{41.10}& \color{black}{-}& \color{black}{-}& \color{black}{-}\\

\color{black}{VLKD~\cite{dai2022enabling}} & \color{black}{42.60}& \color{black}{-}& \color{black}{-}& \color{black}{-}\\

\color{black}{FewVLM~\cite{jin2021good}} & \color{black}{47.70}& \color{black}{-}& \color{black}{-}& \color{black}{-}\\

\color{black}{PNG-VQA$_{3B}$~\cite{tiong2022plug}} & \color{black}{62.10}& \color{black}{-}& \color{black}{-}& \color{black}{-}\\

\color{black}{PNG-VQA$_{11B}$~\cite{tiong2022plug}} & \color{black}{63.30}& \color{black}{-}& \color{black}{-}& \color{black}{-}\\

\color{black}{Img2LLM$_{66B}$~\cite{guo2023images}} & \color{black}{59.90}& \color{black}{-}& \color{black}{-}& \color{black}{-}\\

\color{black}{Img2LLM$_{175B}$~\cite{guo2023images}} & \color{black}{60.60}& \color{black}{-}& \color{black}{-}& \color{black}{-}\\

\color{black}{BLIP-2 ViT-g FlanT5$_{XL}$~\cite{li2023blip2}} & \color{black}{63.10}& \color{black}{-}& \color{black}{-}& \color{black}{-}\\

\color{black}{BLIP-2 ViT-g FlanT5$_{XXL}$~\cite{li2023blip2}} & \color{black}{65.20}& \color{black}{-}& \color{black}{-}& \color{black}{-}\\


DTNet (Ours) & \textbf{67.36}        & \textbf{84.96}       & \textbf{49.38}       & \textbf{58.74}      \\ \bottomrule
\end{tabular}
}
\end{table*}

\begin{figure*}
\vspace{-0.0cm}
\centering 
  \includegraphics[width=1.80\columnwidth]{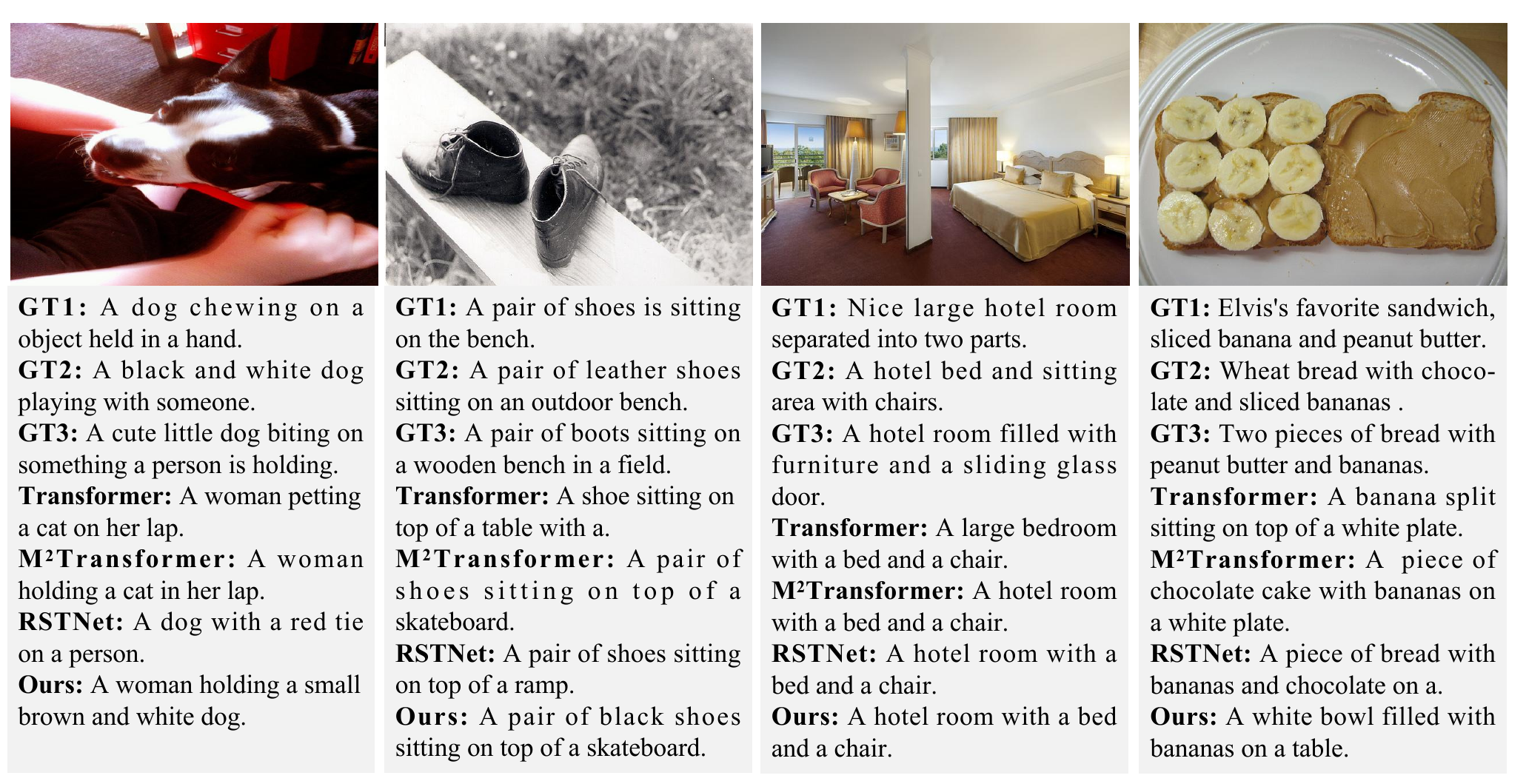}
  \vspace{-0.0cm}
  \caption{ 
     {\color{black}{Negative qualitative visualization obtained by Transformer~\cite{vaswani2017attention}, $M^2$Transformer~\cite{cornia2020meshed}, RSTNet~\cite{zhang2021rstnet} and the proposed method. ``GT'' is short for ``Ground Truth''.}}
  }
  \vspace{-0.0cm}
  \label{fig:badcase}
\end{figure*}

\subsection{Significance Test}

To illustrate the efficacy and superiority of our proposed DTNet, we performed a detailed comparison of the DTNet against the Standard Transformer. Specifically, we conduct a two-tailed t-test with paired sample for each metric to see if the improvement induced by our DTNet is statistically significant. To verify whether the semantics of the caption generated by DTNet is significantly improved over the standard transformer,  we also report semantic subcategories of SPICE scores (\emph{i.e.},  Relation,  Cardinality,  Attribute,  Size,  Color, and Object) of two models. Furthermore, we conducted a two-tailed t-test with paired samples for every detailed SPICE score.

The popular metrics and p-values for t-test are shown in Tab.~\ref{tab:p_metric}. As we can observe, all popular metrics for image captioning are significantly improved under a significant level $\alpha = 0.05$, which proves the effectiveness of our proposed DTNet. Additionally, the detailed SPICE scores and corresponding p-values for t-test are illustrated in Tab. \ref{tab:spice}. We can observe that all the detailed semantic subcategories of SPICE attain improvements. Besides, \emph{Attribute}, \emph{Color} and \emph{Object}  SPICE scores are significantly improved under the significant level $\alpha = 0.05$, which proves that our proposed DTNet can fully mine the semantics in images and generate accurate captions.

{
\color{black}{
\subsection{Generalization On Visual Question Answering (VQA)}
{\color{black}{While the main focus of DTNet is image captioning, we also explore its performance on other multi-modal tasks, such as Visual Question Answering (VQA). To thoroughly evaluate DTNet's capabilities beyond its primary application, we conducted extensive experiments on the widely recognized VQA-V2 dataset. Our findings, presented in Tab.~\ref{tab:vqa}, demonstrate that our proposed DTNet model excels in the VQA task, highlighting its generalizability and versatility. Specifically, when compared to MCAN~\cite{yu2019deep}, a static Transformer-like architecture, our method exhibits substantial improvements across all metrics.}}

}
}

\subsection{Qualitative Analysis}


\subsubsection{Path Analysis}
In Fig. \ref{fig:fig7}, we present a variety of images passing through different number of paths.  Concretely, we employ 0.3 as the threshold to discretize the learned paths (\emph{i.e.,} the paths with the weights less than this threshold are removed). Notably, the number of paths generally increases with the complexity of images increasing, which is compatible with the human perception system \cite{Walther2011SimpleLD}. The reason may be that a small number of cells are enough to handle simple images, and only complex images need the participation of more cells. Fig. \ref{fig:fig8} illustrates customized paths for different images.

\subsubsection{Caption Quality}
Fig.~\ref{fig:fig5} illustrates several image captioning results of Transformer and DTNet for similar images. 
{\color{black}{Significantly, the first two rows of Fig.~\ref{fig:fig5} demonstrate that the Transformer model fails to discern the nuances among similar images, leading it to generate identical descriptions. In contrast, our DTNet exhibits sensitivity to distinguishing the specific characteristics of different samples, allowing it to customize appropriate pathways and generate informative captions. This result once again highlights the superiority of the dynamic scheme employed in our image captioning approach.
To illustrate this further, refer to the first two columns of the first row in Fig.~\ref{fig:fig5}. The Transformer model generates the same caption, ``A bathroom with a toilet and a sink.", for these two distinct images. Conversely, our DTNet accurately discerns the differences in details between the two images and generates distinct descriptions for each.
Furthermore, it is worth noting that the Transformer model may produce incorrect captions to describe images, whereas captions generated by DTNet exhibit higher accuracy. As evident in the first column of the last row in Fig.~\ref{fig:fig5}, we observe that the Transformer model mistakenly predicts the number of trains, resulting in an incorrect caption, ``Two red trains are on the tracks in the snow." Conversely, our proposed model generates a precise caption, ``A red train is on the tracks in the snow."
}}
To gain deep insights into each path of our DTNet, we randomly sample four paths and illustrate the generated captions of these sampled paths in Fig. \ref{fig:fig6}. An interesting observation is that the captions generated by different paths are diverse yet accurate. Therefore,  in addition to achieving new state-of-the-art performance, our DTNet also provides a new approach for DIV.

{
\color{black}{
\subsubsection{Limitations}
While our proposed DTNet has demonstrated exceptional performance, it is important to acknowledge its limitations. Firstly, DTNet may occasionally make incorrect predictions for objects that share similar appearances. For instance, as depicted in the first column of Fig.~\ref{fig:badcase}, a black and white dog may exhibit fur colors that appear similar to brown and white under sunlight. Consequently, our DTNet may mistakenly predict it as a brown and white dog.
Additionally, in complex scenes, DTNet may struggle to capture and describe all the intricate details present. This is evident in the third column of Fig.~\ref{fig:badcase}, where DTNet generates the caption ``A hotel room with a bed and a chair" to describe the image. Although the generated caption is error-free, it fails to provide an exhaustive description of all the details within the image.
}
}


\section{Conclusion}

In this paper, we present \emph{Dynamic Transformer Network} (DTNet)  for image captioning. Concretely, we introduce five basic cells to construct the routing space, and group them by domains to achieve better routing efficiency. We propose the \emph{Spatial-Channel Joint Router} (SCJR) for customizing dynamic paths based on both spatial and channel information of inputs. Extensive results on the MS-COCO benchmark demonstrate the superiority of our proposed DTNet over previous SOTAs. The presented cell design and routing scheme also provide insights for the future study of input-sensitive learning methods.


%



\ifCLASSOPTIONcaptionsoff
  \newpage
\fi



%

%
%

\bibliographystyle{./IEEEtran}
\bibliography{./IEEEabrv,./IEEEexample}

\end{document}